\pdfoutput=1
\documentclass[11pt]{article}

\usepackage[]{acl}

\usepackage{times}
\usepackage{latexsym}
\usepackage{booktabs}

\usepackage[T1]{fontenc}
\usepackage[utf8]{inputenc}

\usepackage{microtype}

\usepackage{inconsolata}

\usepackage{graphicx}

\usepackage{tikz}

\usepackage{colortbl}
\usepackage{array}
\usepackage{caption}
\usepackage{pifont}
\definecolor{user2color}{rgb}{1, 0.8, 0.8} % Light red
\definecolor{user1color}{rgb}{0.8, 1, 0.8} % Light green
\definecolor{user3color}{rgb}{0.8, 0.8, 1} % Light green
\definecolor{user4color}{rgb}{1.0, 0.85, 0.65} % Light orange
\definecolor{lightgrey}{rgb}{0.9, 0.9, 0.9} % Light grey

\title{An LLM Feature-based Framework\\for Dialogue Constructiveness Assessment}

\usepackage{authblk}

\author[1]{Lexin Zhou}
\author[1,2]{Youmna Farag}
\author[1]{Andreas Vlachos}

\affil[1]{University of Cambridge}
\affil[2]{Toshiba Cambridge Research Laboratory}

\affil[ ]{\texttt{lz473@cam.ac.uk, Youmna.Farag@toshiba.eu, av308@cam.ac.uk}}

\begin{document}
\maketitle
\begin{abstract}
Research on dialogue constructiveness assessment focuses on (i) analysing conversational factors that influence individuals to take specific actions, win debates, change their perspectives or broaden their open-mindedness and (ii) predicting constructiveness outcomes following dialogues for such use cases. These objectives can be achieved by training either interpretable feature-based models (which often involve costly human annotations) or neural models such as pre-trained language models (which have empirically shown higher task accuracy but lack interpretability). In this paper we propose an LLM feature-based framework for dialogue constructiveness assessment that combines the strengths of feature-based and neural approaches, while mitigating their downsides. The framework first defines a set of dataset-independent and interpretable linguistic features, which can be extracted by both prompting an LLM and simple heuristics. Such features are then used to train LLM feature-based models. We apply this framework to three datasets of dialogue constructiveness and find that our LLM feature-based models outperform or performs at least as well as standard feature-based models and neural models.  We also find that the LLM feature-based model learns more robust prediction rules instead of relying on superficial shortcuts, which often trouble neural models.%Further, interpreting these LLM feature-based models yield numerous valuable insights into what makes a dialogue constructive.
\footnote{The experimental code and data can be found in our GitHub repository: \url{https://github.com/lexzhou/llm-feature-based-framework-for-DCA}.}
\end{abstract}

\section{Introduction}

\begin{figure}[h]
    \centering
    \includegraphics[width=0.47\textwidth]{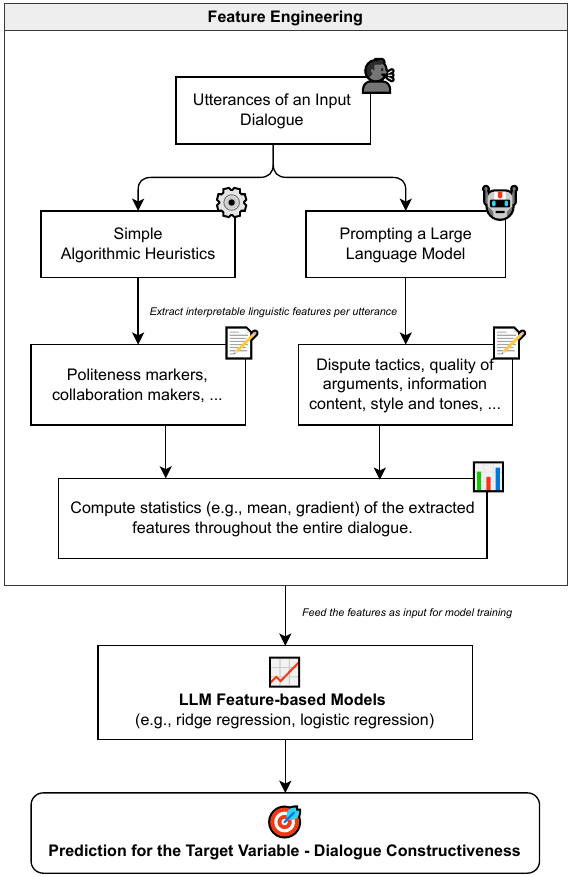}
    \caption{Flowchart delineating a high-level overview of our proposed framework for dialogue constructiveness assessment.}
    \label{fig:abstract_graph}
\end{figure}

Numerous prior studies analyse or forecast the constructiveness of conversations spanning a wide range of topics. These include the examination of dialogues aimed at persuading individuals to take specific actions like contributing to charitable causes  \citep{wang2019persuasion, shi2020effects}, 
shifting their stance \citep{tan2016winning, prakken2020persuasive}, garnering more votes \citep{zhang2016conversational, slonim2021autonomous}, argumentation to achieve consensus \citep{mayfield2019analyzing, vecchi2021towards, de2021beg, de2022disagree}, and investigating how dialogues foster open-minded thinking \citep{farag2022opening}. Two main goals of these studies are to (i) better understand the interplay between linguistic patterns humans exhibit and the targeted outcomes associated with dialogue constructiveness and (ii) build models that can anticipate the outcome of constructive dialogues for certain use cases.

These goals are usually achieved by building two types of models: Feature-based models, which take linguistic features as input to predict the targeted outcome, and neural models like Pre-trained Language Models (PLM), which take adialogue (or a part of it) as input. The former is usually an interpretable %\footnote{Assuming the input dimension or the number of input variables that matter is reasonably manageable for human interpretation.} 
model (e.g., logistic regression), while the latter can be artificial neural networks like BERT \citep{devlin2018bert}, Longformer \citep{beltagy2020longformer} and few-shot decoder-only transformers \citep{achiam2023gpt} that tend to produce higher task accuracy.

However, these two types of models have downsides. For example, the input data of feature-based models often require expensive human annotations (e.g., levels of disagreements and coordination tactics) \citep{meyers2018dataset, wang2019persuasion, de2022disagree}, whilst neural models are black-boxes. Further, these models are susceptible to learning shortcuts—decision rules that perform well for the majority of benchmark examples but do not hold in general \citep{gururangan2018annotation, geirhos2020shortcut}. For instance, neural models are prone to learning shortcuts\footnote{We use the terminology introduced by \citet{geirhos2020shortcut}. Other works also refer to shortcuts as “biases” or “spurious correlations”.} due to the exposure to a rich input space% from natural textual data, allowing PLMs to exploit various patterns for inference, including superficial ones
. Likewise, feature-based models can learn shortcuts, though this issue can be mitigated by careful feature engineering.

To our knowledge, prior works use distinct sets of features without a unifying framework, many of which were tailored to a specific context (e.g., specific persuasion tactics for soliciting donations).  This work contributes by introducing a unifying framework for dialogue constructiveness assessment (Figure \ref{fig:abstract_graph}) that capitalises on the strengths of both model types while mitigaiting their shortcomings. It integrates both simple algorithmic rules and prompting a Large Language Model (LLM), namely GPT-4 \citep{achiam2023gpt}, to automatically extract a rich set of dataset-independent\footnote{Namely, these features that our framework provides are applicable to any dataset related to dialogue constructiveness and potentially adaptable to other dialogue-style datasets.} linguistic features, such as the average occurrence and usage change of collaboration markers \citep{niculae2016conversational} %, politeness markers \citep{zhang2018conversations}, 
and dispute tactics \citep{de2022disagree} used throughout a given dialogue. We will refer to the features generated by heuristics and LLMs as \textit{discrete} and \textit{LLM-generated} features, respectively. These features are then fed into ridge/logistic regression algorithms, which are referred to as \textit{LLM feature-based models}.  We find that the patterns learned by our LLM feature-based models have better predictive power and robustness than standard feature-based and neural models, and showcase our framework's ability to spot predictive linguistic factors for the targeted outcomes across three datasets of dialogue constructiveness: Opening-up Minds (OUM) \citep{farag2022opening}, Wikitactics \citep{de2022disagree} and Articles for Deletion (AFD) \citep{mayfield2019analyzing}. Table \ref{tab:three_examples} shows three examples (one per dataset) with dialogue segments, target variables, and subsets of extracted features.

\begin{table*}[h!]
\centering
\scriptsize
\begin{tabular}{|p{6.5cm}|p{4.6cm}|p{3.5cm}|}
\hline
\textbf{Dialogue} & \textbf{Target Variable} & \textbf{Linguistic Features} \\ \hline
\begin{tabular}[c]{@{}p{6.5cm}@{}}
\cellcolor{user1color}{\tt$<$WoZ$>$ Hello! Would you like to tell me if you think the COVID-19 vaccination is a good idea?} \\
\cellcolor{user4color}{\tt$<$participant$>$ It's a good idea as Covid has taken so many lives and disrupted so many businesses and livelihoods. So having a vaccine that can prevent people from dying is good and we can be safe in public again knowing the chances of dying are low.} \\
\cellcolor{user1color}{\tt$<$WoZ$>$ That’s a fair point! However, some people are concerned about the vaccines because there have been recorded cases of people dying from blood clots after taking them, and others have been seriously ill with side-effects.} \\
\cellcolor{lightgrey}{\tt......} \\
%$<$participant$>$ I don't know the exact numbers but I think its low from the people who actually died from Covid. I think the risk is worth it. \\
%$<$WoZ$>$ True, though others argue that the vaccine is unnecessary and the long term effect is unknown at this stage. Especially if they are healthy and under 70, the risk posed by COVID to them is so small as to be not worth taking the vaccine. \\
%$<$participant$>$ Fair point, I however think the long terms of contracting Covid are also unknown. We've seen healthy young people die from it. I don't think taking the vaccine is unnecessary and the sooner we get back to normal life where being out in public is not as deadly the better. \\
%$<$WoZ$>$ That is a fair point! However, some would argue that the need to protect others through herd immunity does not justify mandating the vaccine for those who are opposed to getting it. \\
\cellcolor{user4color}{\tt$<$participant$>$ We were mandated to go under strict lockdown, with no choices. I think laws from companies and establishments making vaccinations compulsory is great in achieving herd immunity and those who don't comply must stay away from society.} \\
\cellcolor{user1color}{\tt$<$WoZ$>$ Some are concerned that mandatory vaccination takes away an individual's right to choose with informed consent, and without informed consent it is medically unethical to force a medicine, medical procedure, or surgery upon the patient.}
\end{tabular} &
\begin{tabular}[c]{@{}p{4.6cm}@{}}
{\tt Post-conversation open-mindedness score measuring the participant's belief that those who hold opposing viewpoints have good reasons for their beliefs (on a 7-point Likert scale) after having a conversation on a given controversial topic (e.g., COVID-19 vaccination) with another human or a chatbot. The post-conversation open-mindedness score from the participant for this instance is 3.}
\end{tabular} &
\begin{tabular}[c]{@{}p{3.5cm}@{}}
{\tt 1st person pronouns, $\bar{x}$=0.22}\\ \\
{\tt Hedge words, $\bar{x}$=0.67}\\ \\
{\tt Gratitude, $\bar{x}$=0.00}\\ \\
{\tt \# words, $\bar{x}$=35.78}\\ \\
{\tt Attempted derailing/off-topic, $\bar{x}$=0} \\\\
{\tt Asking questions, $\bar{x}$=0.11}\\ \\
{\tt Counterargument, $\nabla$=0.067}\\ \\
{\tt Suggesting a compromise, $\nabla$=0}\\ \\
{\tt Neutral sentiment, $\bar{x}$=0.33}\\ \\
{\tt High formality, $\bar{x}$=0.67} \\ \\
%{\tt High content density, $\nabla$=0}\\ \\
{\tt Epistemic uncertainty, $\nabla$=0}\\ \\
\end{tabular} \\ \hline
\begin{tabular}[c]{@{}p{6.5cm}@{}}
\cellcolor{user4color}{\tt$<$user\_id=Ilovetopaint$>$ You know very well that the track was named in honor of Sean's band, and that mentioning it genuinely informs the context of the album. What is with these pedantic reverts?}\\
\cellcolor{user1color}{\tt$<$user\_id=153.205.69.164$>$ Could you provide the source that states "'Micro Disneycal World Tour' was named in honor of Microdisney"?}\\
\cellcolor{user4color}{\tt$<$user\_id=Ilovetopaint$>$ This has already been satisfied by two references which observe the fact that a guy who was in a band called "Microdisney" remixed the album track called "Micro Disneycal World Tour". Even if the song wasn't actually named after the band (yeah right), the fact that a member of ''that'' band happened to remix ''that'' track would still be of encyclopedic interest.}\\
\cellcolor{lightgrey}{\tt......} \\
\cellcolor{user1color}{\tt$<$user\_id=153.205.69.164$>$ Besides, I have no idea what makes you want to mention only Sean O'Hagan in the article. As credited as [...] "The Micro Disneycal World Tour" was remixed by the High Llamas, not only Sean O'Hagan. Even the MTV source you added states that it is "the High Llamas' take on 'The Micro Disneycal World Tour'". He is not the remixer for "Micro Disneycal World Tour". Although Sean O'Hagan is a member of the High Llamas, the High Llamas is not Sean O'Hagan's one-man band.} \\
\end{tabular} &
\begin{tabular}[c]{@{}p{4.6cm}@{}}

{\tt Binary indicator noting whether a suggested Wikipedia page edit, after holding a dispute between several users/editors, would require escalation (i.e., Disputes that the editors are unable to settle on their own are escalated to mediation. This outcome can be regarded as a proxy to dialogue constructiveness.). The outcome for this instance is 1 (i.e., escalated).}

\end{tabular} &
\begin{tabular}[c]{@{}p{3.5cm}@{}}
{\tt 2nd person pronouns, $\bar{x}$=0.36}\\ \\
{\tt Subjunctive words, $\bar{x}$=0.09}\\ \\
{\tt Assert factuality, $\nabla$=-0.03}\\ \\
{\tt Name calling or hostility, $\bar{x}$=0.0}\\ \\
{\tt Counterargument, $\nabla$=0.05}\\ \\
{\tt Providing clarification, $\bar{x}$=0.1}\\ \\
{\tt No uncertainty, $\nabla$=0.08}\\ \\
%{\tt Low formality, $\bar{x}$=0.2}\\ \\
{\tt Negative sentiment, $\nabla$=0.06}\\ \\
{\tt High politeness, $\nabla$=-0.06}\\ \\
\end{tabular} \\ \hline
\begin{tabular}[c]{@{}p{6.5cm}@{}}
\cellcolor{user3color}{\tt$<$user\_name=Eastmain$>$ [[User:Ulflarsen]] prodded with the comment: "A near identic article on the Norwegian bokmål/riksmålswikipedia were recently removed as the major Norwegian newspaper VG in a mail claimed that the article were a violation of it's right to the material, it has been compiled from that, so the article should be deleted here as well" and then [[User:Nsaa]] added an AfD notice. Technical nomination only. I cannot tell if this is a copyvio or not.}  \\
\cellcolor{user4color}{\tt$<$user\_name=Nsaa$>$ * VOTE It has been deleted on Norwegian Wikipedia [...], maybe because the list contains the publishers name in the article name [[VG]] (this is not the case here). This list can be compared to [[Pop 100 number-one hits of 2007 (USA)]]  $<$small$>$—Preceding [[Wikipedia:Signatures|unsigned]] comment added by.}\\ 
% $<$user\_name=Michael A. White$>$ *VOTE. They don't have a copyright on facts.-- \\
\cellcolor{lightgrey}{\tt......} \\
\cellcolor{user2color}{\tt $<$user\_name=Chubbles$>$ *VOTE clearly encyclopedic, and a list of hits is not a copyright violation. The unlinked artists should be linked, as they are all obviously notable.} \\ 
\cellcolor{user1color}{\tt $<$user\_name=Mm40$>$ *VOTE As long as everything is legal and all, it should be kept. Notable and not bad enough to warrant a deletion.}\\
\end{tabular} &
\begin{tabular}[c]{@{}p{4.6cm}@{}}
{\tt Binary indicator representing whether a nominated article would be kept or deleted by Wikipedia administrators after the dispute is held between Wikipedia users in which any user, including those who are unregistered but sign their post with an IP address, can cast a vote and provide a rationale for their stance on whether the article should be retained or removed. Users may also contribute non-voting comments. The outcome for this instance is 1 (i.e., kept).}
\end{tabular} &
\begin{tabular}[c]{@{}p{3.5cm}@{}}
{\tt Start with greeting, $\nabla$=0}\\ \\
{\tt \# 3rd person pronouns, $\bar{x}$=0.66}\\ \\
{\tt Certainty terms, $\nabla$=0.06}\\ \\
{\tt Refutation, $\bar{x}$=0.27}\\ \\
{\tt Stating your stance/repeated argument, $\bar{x}$=0.73}\\ \\
{\tt Providing clarification, $\bar{x}$=0.27}\\ \\
{\tt Positive sentiment, $\nabla$=0}\\ \\
% {\tt High propositional density, $\nabla$=-0.02}\\ \\
{\tt Low formality, $\bar{x}$=0.36}\\ \\
{\tt High politeness, $\nabla$=0.02}\\ \\
\end{tabular} \\ \hline
\end{tabular}
\caption{Three example dialogue segments from the three datasets included for our analysis in tandem with their target variables and nine examples of extracted linguistic features (averaged across utterances, $\bar{x}$, or the slope of a linear fit, $\nabla$) generated with simple heuristics and GPT-4. From top to bottom, these examples belong to OUM, Wikitactics and AFD datasets. Details of the features and the datasets are described in section \ref{sec:framework_definition} and \ref{sec:modelling_and_evaluation}, respectively. The utterances of distinct entities are coloured differently. The example from OUM, in the first row, `WoZ' refers to a human pretending to be a dialogue agent via an appropriate interface in the context of a Wizard-of-Oz experiment.
}
\label{tab:three_examples}
\end{table*}

\section{Background and Related Work}

\subsection{Dialogue Constructiveness}

Analysing dialogue constructiveness has long been a research field of important interest. In recent years, it has become increasingly more relevant as more and more conversations online are not task-focused/customer services-related but more open-ended. Prior work has focused on 1) understanding the factors that can impact and 2) building models that can predict dialogue constructiveness%(e.g., problem-solving, group decision, resolving disagreements)
. For instance, \citet{zhang2016conversational}  track how ideas flow between participants throughout a debate%, i.e., analysing how debaters strategically use and respond to both pre-planned talking points and emergent discussion points as the conversation develops, in a case study of Oxford-style debates (in which winners are chosen based on audience votes)
. Statistical analysis reveals that victorious debaters more effectively leverage the interactive aspect of the debate by engaging with their opponents' arguments.
%throughout the discussion instead of focusing on advancing their own viewpoints.

\citet{mayfield2019} perform an analysis on group decision-making on the `Articles for Deletion' Wikipedia forum, where editors deliberate on whether an article should be removed% from the site
. %In these disputes, users may participate in the conversation and vote on whether to keep or remove nominated articles. 
The authors train a BERT model to forecast the outcomes of these debates based on the discussion text and measure the impact of discursive strategies and different user behaviours on the predicted outcomes %(e.g., individuals who contributed more frequently were less likely to end up winning) 
by monitoring the shift in the model's predicted probabilities after each contribution to the discussion. %Their model revealed that (1) early votes are strongly predictive of final decisions; (2) citing specific Wikipedia policies and guidelines is associated with both forecast shift and success in swaying the outcome; (3) individuals who contributed more frequently were less likely to end up on the winning side, although their initial votes tended to have a greater impact on the model's forecasts compared to one-time participants; and (4) votes cast later in the discussion had minimal effect on altering the anticipated outcome, suggesting that the deliberation had essentially concluded by that stage.

\citet{wang2019persuasion} collect a dataset of dialogues where one person persuades the other to donate to charity. %They designed a persuasion task where pairs of participants engaged in a conversation, with one participant (the persuader) attempting to convince the other (the persuadee) to donate to the charity ``Save the Children''. 
%They annotate a subset of the collected conversations with ten persuasion strategies used by the persuader, such as emotional appeal, personal story, and donation information. With such annotations, 
They train a feature-based model (using logistic regression) to examine the associations between the human-annotated persuasion strategies and the binary donation outcome (1 = donation, 0 = no donation). They find that only the strategy of providing information about the donation procedure has a significant positive effect on donation.% In addition, they analysed the relationships between participants' psychological and demographic backgrounds and donation decisions by interpreting the coefficients of the feature-based model. %They found that certain traits, such as agreeableness, were associated with higher donation probability. Interestingly, they also discovered interaction effects between personal characteristics and the effectiveness of different persuasion strategies. For instance, the source-related inquiry strategy (asking about familiarity with the charity) was more effective for individuals high in openness.

\citet{de2021beg} analyse factors that make dispute resolution work on Wikipedia Talk Page by training feature-based models % (fed with linguistic features like collaboration and politeness markers, toxicity score and sentiment) 
to predict escalation\footnote{Disputes which cannot be resolved by editors are escalated to mediation, which is a proxy for constructiveness.}. They find both average occurrence and change in the use of collaboration and politeness markers are predictive of escalation. 
%and that the model performance can be enhanced not only by considering the average occurrence of linguistic markers but also by features that capture changes in linguistic markers
In addition, they develop neural models and show that accounting for the conversation's structure improves predictive accuracy, outperforming feature-based models.%, suggesting that neural models may be a more desirable option for forecasting the targeted outcome of dialogue constructiveness of online deliberation

%Similarly, \citet{de2022disagree} introduce a framework of dispute tactics that takes into account the use of different rebuttal strategies and coordination tactics. They train a hierarchical attention network and demonstrate that human-annotated utterance-level features, such as the average occurrence and change of disagreement strategies used throughout the conversation, can help predict dispute escalation. % Removed to reduce words, since we mention it later on section 4.

\citet{farag2022opening} focus on argumentative dialogues that aim to foster open-mindedness towards views that one opposes. They introduce a dataset of 183 human-human and human-chatbot dialogues on three controversial topics: Brexit, veganism, and COVID-19 vaccination. %The human-human dialogues were collected using a Wizard of Oz (WoZ) approach (i.e. a moderated research method in which a user interacts with an interface that appears to be autonomous but is actually controlled by a human). The wizards and the chatbot used arguments from a knowledge base to converse with participants. Before the discussion, the participant would be asked about if they had a preference for one particular stance in an implicit way (e.g., if they are vegans or not) while the other side does not hold any particular stance. 
To measure open-mindedness, the authors ask participants before and after the dialogue whether they think people with opposing views have good reasons for their stance, intellectual capabilities, and morality. %The success of a dialogue in opening up minds was measured by the change in participants' responses to these questions. The authors assess two dialogue models. One is Wikipedia-based and another argument-based. They found both perform comparably in opening up minds. 
They find that participants become more open to the reasons behind opposing views but not so regarding the morality and intellectual capabilities of their opponents. There is no strong correlation between chat experience ratings (e.g., clarity, enjoyment, persuasiveness) and changes in open-mindedness% nor between the average use of politeness features and the success of dialogue in opening up minds
.

\citet{khan2024debating} investigate the use of debate between two LLM experts each arguing for a different answer on a reading comprehension task, as a method to elicit truthful information, and ask a (human/LLM) non-expert to judge the debate and select the answer. They find that debate protocols outperform the single-model consultancy approach. They also show that the level of persuasiveness of debaters can have a positive impact on the accuracy of judgements of non-experts, implying that optimising features like the level of persuasiveness during the debate scenario may enhance dialogue constructiveness and promote the user's ease in the verification of correct output or errors.

\subsection{Shortcut Learning}

%The remarkable success of deep learning models for the last decade may have somehow overshadowed the necessity for reliable explanations of their decision-making processes. 
Neural models have a tendency toward learning shortcuts \citep{geirhos2020shortcut, du2023shortcut}, despite their impressive performance. For instance, imperceptible perturbations to the input \citep{szegedy2013intriguing} or changes in the background context \citep{beery2018recognition, rosenfeld2018elephant} can completely derail their predictions.
Also, neural models could do `natural language inference' by just detecting correlated keywords instead of engaging in genuine reasoning \citep{gururangan2018annotation}. %Further, \citet{si2022spurious} has reported systematic patterns when fine-tuning BERT and few-shot prompting GPT-3. They exploit certain types of spurious features (e.g., content words) to a significantly larger extent than others (e.g., function words). 

Neural dialogue evaluation metrics can also exhibit shortcut behaviour or biases, as demonstrated by \citet{khalid2022explaining} through an adversarial test suite that revealed these metrics often fail to properly penalize problematic conversations and may rely on superficial patterns rather than deep semantic understanding. Further, LLMs may learn shortcuts or brittle representations \citep{zhang2024comprehensive, chen2024unveiling}, thus exhibiting inconsistent performance across various adversarial perturbations designed to degrade dialogue quality%, suggesting that LLMs may be using shortcuts that fail to capture these aspects of dialogue quality. Similarly
. LLMs may also exploit shortcuts in prompts during in-context learning \citep{tang2023large}, raising concerns about robustness and generalisability.% Larger models, in particular, were found more likely to utilise these shortcuts%, posing significant challenges in ensuring their reliability across diverse contexts

Given these observations, we hypothesise that the neural models we develop to predict dialogue constructiveness may learn shortcuts% rather than capturing the underlying linguistic patterns that contribute to constructive conversations
. Nevertheless, feature-based models can also learn shortcuts, though the risk can be mitigated by carefully crafting features that capture patterns that generalise well across datasets.%, in which we make an analysis and comparison with respect to the extent to which shortcut learning occurs in both neural and feature-based models for dialogue constructiveness prediction tasks

\section{LLM Feature-based Framework}
\label{sec:framework_definition}

\begin{table*}[h!]
    \centering
    \scriptsize
    \begin{tabular}{|p{2.2cm}|p{12.8cm}|}
    \hline
        \textbf{Feature Set} & \textbf{Description} \\\hline
        
        Politeness markers & The politeness strategies introduced by \citet{zhang2018conversations} and implemented in Convokit \citep{chang2020convokit}, encompassing greetings, apologies, directness levels, and the use of polite expressions. \\\hline
        
        Collaboration markers & Conversation markers that are indicative of collaborative discussions \citep{niculae2016conversational}, including the introduction and adoption of ideas, expressions of uncertainty or confidence, pronoun usage patterns, and linguistic style accommodation. \\\hline
        
        Dispute tactics & Indicators concerning the use of distinct rebuttal strategies (i.e., attempts to counter the arguments of an opponent)
        and coordination tactics (i.e., attempts to promote understanding and consensus), introduced in \citep{de2022disagree}. \\\hline
        
        Quality of Arguments (QoA) & Assessing the QoA has been an important area of research in linguistics and NLP \citep{lawrence2020argument}. Despite the subjectivity, it can be generally agreed that arguments based on scientific evidence are more convincing than those based solely on speculation from an individual without expertise on the subject matter. %In this work, we hypothesise that the QoA may provide predictive power for dialogue constructiveness. 
        We explore a new way to model argument quality: Prompting an LLM to grade the quality of all arguments used in a discussion on a 0-10 scale. \\\hline
        
        Information content & This includes content density and propositional density. Content density is the ratio of open-class to closed-class words, while propositional density measures the ratio of propositions to the number of words \citep{roark2011spoken}. \\\hline
        
        Style and tone & This set contains several features characterising the style and tone of utterances. SENTIMENT: To capture the emotional tone, utterances are classified as negative, neutral, or positive. POLITENESS: This feature evaluates the degree of politeness expressed in the text, classified as high or low politeness. FORMALITY: To capture speech formality, distinguishing between casual remarks and professionally structured utterances. UNCERTAINTY: Individuals can express varying degrees of uncertainty during a discussion, which might be predictive of dialogue constructiveness. We annotate four types of uncertainty (i.e., epistemic, doxastic, investigative, conditional) as in \citep{meyers2018dataset}. \\\hline
        
    \end{tabular}
    \caption{Six dataset-independent feature sets. The top two are discrete features generated using heuristics, while the other four are LLM-generated. Details on individual features in each set are in Appendix \ref{app:feature_descriptions}. The prompts used for LLM-generated features are in Appendix \ref{app:prompts_generating_feats}. Examples of extracted features are shown in Table \ref{tab:three_examples}.}
    \label{tab:features}
\end{table*}

The LLM feature-based framework that we introduce works as follows: Given a labelled dataset of dialogue constructiveness, we generate linguistic features at (i) the \textit{utterance-level} for all utterances or (ii) at the \textit{dialogue-level}, for every input dialogue. More concretely, we include six linguistic feature sets (Table \ref{tab:features}) in which five are sourced from the literature and one (i.e., QoA) is newly introduced in this work; all feature sets are annotated at the utterance-level except for the QoA, which is generated at the dialogue-level. These features are all dataset-independent and are annotated using either heuristics or prompting an LLM with in-context learning.\footnote{Specifically, we prompt GPT-4-1106-preview via OpenAI API, which is referred to as GPT-4 in this work. We also explored two frontier open-source LLMs (LLaMA-3.1-70B-Instruct and LLaMA-3.1-405B-Instruct), but they exhibited worse annotation quality  (see Appendix \ref{ap:llama3dot1_annot_quality}).} %Noteworthy, 
We note that the quality of annotations of the used LLM should be validated manually, e.g., a human expert taking a sample of annotated features and inspecting their annotation accuracy.

Once the linguistic features are generated and validated, we compute statistics for each aforementioned feature annotated at the utterance-level (i.e., all features sets except for QoA) to capture the dynamics of dialogue; the resultant statistics would be considered as dialogue-level features. In particular, we compute the \textit{average} and the \textit{gradient} (slope of a linear fit) statistics to the feature values of all utterances throughout the entire dialogue, as proposed by \citet{de2021beg}. 

Lastly, we can train any interpretable regression or classification algorithm (e.g., ridge/logistic regression) by feeding the computed statistics or dialogue-level features to the model, to predict the target variable of interest. Figure~\ref{fig:abstract_graph} summarises the framework. Noteworthy, the framework can be used flexibly, e.g., one may include dataset-specific features to further improve the LLM feature-based model's performance. For instance, instead of calculating statistics for each feature's values throughout the \textit{entire} dialogue, one can also explore the option of disaggregating them at the \textit{participant level}.\footnote{The statistics of each feature are computed from each participant's utterances instead of from the entire conversation.} As we will demonstrate in our experimental results, this setup is readily applicable to one dataset that we analyse, where there is always a fixed number of speakers throughout the conversation. This setup may also be applied to datasets where the number of individuals varies, but it will require further adaptations.

\section{Experimental Setup}
\label{sec:modelling_and_evaluation}

\subsection{Data}
\label{sec:data}

We analyse three datasets (in English) related to dialogue constructiveness assessment to showcase the utility of the proposed framework: OUM \citep{farag2022opening}, Wikitactics \citep{de2022disagree}, and AFD \citep{mayfield2019analyzing} corpus. Table \ref{tab:dataset_stats} provides statistics of the dialogues in these corpora. 

\begin{table*}[h!]
\footnotesize
\centering
\begin{tabular}{@{}lccc@{}}
\toprule
 & OUM & Wikitactics & AFD \\ \midrule
\# Dialogues & 542 & 213 & 1000 \\ 
Avg. \# Utterances per dialogue & 26.0 (12.4) & 19.3 (7.8) & 9.7 (13.6) \\ 
Avg. \# Words per utterance & 27.0 (20.9) & 70.2 (74.3) & 32.1 (41.5) \\ \bottomrule
\end{tabular}
\caption{Dialogue Statistics (mean and standard deviation) for the three datasets of our analysis.}
\label{tab:dataset_stats}
\end{table*}

\paragraph{OUM} 542\footnote{There were 183 dialogues on the original paper and later extended by 359 additional human-chatbot dialogues. Among these, 99 are human-human, and 443 are human-chatbot.} argumentative dialogues between participants and chatbots/WoZ (i.e., the Wizard of Oz method, a moderated research method in which a user interacts with an interface that appears to be autonomous but is actually controlled by a human) on controversial topics \citep{farag2022opening}. The topics include Brexit ($n=149$), COVID-19 vaccination ($n=189$) and veganism ($n=204$). The dialogues aim to facilitate open-mindedness by exposing participants to opposing viewpoints, with arguments sourced from an online debate platform, Kialo\footnote{https://www.kialo-edu.com/}. Following each dialogue, participants rated their post-conversation open-mindedness towards the opposing stance on a 7-point Likert scale, assessing whether they believe people with opposing views to theirs have good reasons. We model this as a proxy for dialogue constructiveness as a regression task.

\paragraph{Wikitactics} 213 disputes sourced from the Wikipedia Talk Page \citep{de2022disagree}. When there is a content accuracy dispute or a violation of Wikipedia's neutral point of view policy, an editor can create a `dispute' for a potentially problematic article, in which they provide their rationale, vote and discuss them with others. If the editors cannot reach an agreement, they can request mediation from a community volunteer, which is considered an escalation. We model a binary classification task by taking a dialogue as the input to predict two constructiveness outcomes (`escalated' or `non-escalated').

\paragraph{AFD} A sample of 1000 instances randomly extracted from the original 400K discussions (to reduce the excessive use of compute and environmental impact), where editors debate whether to delete certain Wikipedia articles \citep{mayfield2019analyzing}. The corpus was collected by extracting debates (that last at least seven days) initiated by editors, who provide reasoning for nominating an article for deletion. Administrators then aggregate the discussion to decide the outcome, typically following the majority opinion unless a clear consensus is lacking, in which case the article is retained by default. For our analysis, we model a binary classification task by including articles labelled as `Delete' (y=0) and `Keep' (y=1), as per \citet{mayfield2019analyzing}.

\subsection{LLM Feature-based Models}

For the LLM feature-based models, we use ridge regression for OUM and logistic regression for Wikitactics and AFD. In order to assess the annotation quality of `dispute tactics', `information content' and `style and tone' feature sets, one of the authors inspected 100 utterances and found that the accuracy\footnote{`Accuracy' since these feature sets are dichotomised.} ranged between 86\%-100\% % (average 95\%) % removed for cutting spaces.
across individual features, which we deem as acceptably low error rates. Similarly, we check the annotations of `QoA' by extracting 100 pairs of dialogues to conduct pairwise comparison (i.e., \textit{which dialogue should be ranked higher in terms of the QoA?}) to obtain human judgement data. This results in an agreement rate of 81\% between human judgement and LLM's judgement\footnote{This corresponds with a function post-processing the LLM's annotations on the QoA, denoting if a given dialogue A was scored higher than dialogue B. %For instance, if dialogue A and B were annotated with scores 4 and 7, the function would return 0.
}. We consider this agreement rate to be sufficiently high, given this feature's subjective nature. In Appendix \ref{ap:error_analysis}, we provide an error analysis to better understand the noise introduced by the LLM's annotations.

Lastly, for the OUM dataset, in addition to calculating statistics for each feature's values throughout the \textit{entire} dialogue, we also explore the option of disaggregating them at the \textit{participant level} since the number ($N=2$) of speakers is fixed.

\subsection{Baselines}

We include eight well-established baselines: 1) Random classifier or Average regressor; 2) Bag-of-Words, using ridge/logistic regression algorithms; 3) GloVe Embeddings, by feeding the average pre-trained GloVe embeddings \citep{pennington2014glove} as input for ridge/logistic regression algorithms; 4) Longformer (full), which corresponds with fine-tuning the entire \textit{Longformer-large} \citep{beltagy2020longformer}; 5) Longformer (last layer), which fine-tunes only the last layer of \textit{Longformer-large}; 6) zero-shot prompted GPT-4o; 7) twenty-shot prompted GPT-4o\footnote{Appendix \ref{ap:gpt4o_scaling_analysis} describes the used prompts and a scaling analysis that evaluates the performance of N-shot GPT-4o variants, with N $\in \{0, 1, 2, 5, 10, 20\}$.} %We find that zero-shot is the best (and discuss why), and thus further increasing the number of shots is costly and likely to be futile and wasteful.}; 
and 8) Standard feature-based models, fed with discrete features exclusively, using ridge/logistic regression algorithms.

\section{Modelling and Evaluation}

We train all models (except for N-shot GPT-4o baselines) with seven-fold flat cross-validation \citep{wainer2021nested}% for hyperparameter optimisation and evaluation. %This approach is chosen because flat cross-validation provides a balance between computational efficiency and performance assessment reliability and is especially useful for moderately sized datasets like the ones analysed in this work. 
%In contrast to nested cross-validation (which has an inner loop for hyperparameter tuning and another outer loop for performance estimation), flat cross-validation only uses a single cross-validation step to perform both tasks
. 
This method can provide a reliable assessment of model performance while drastically lowering the computational load. It is especially useful for moderately sized datasets like ours and can produce results similar to nested cross-validation for models with relatively few hyperparameters. Appendix \ref{app:model_hyperparam} details the hyperparameter-tuning, training budget and computing infrastructure.

For the evaluation of classifiers, we use the Area Under the Receiver Operating Characteristic curve (AUROC) and the Area Under the Precision-Recall curve (AUPR). For regressors, we use Spearman's Rank Correlation and Mean Absolute Error (MAE). 

We report the average and standard deviation of the results from three seeds for all models except for the 20-shot prompted GPT-4o baseline to due to its cost.\footnote{As noted in Appendix \ref{ap:gpt4o_scaling_analysis}, the 20-shot variant performs considerably worse than the 0-shot variant.} Feature-based models were trained using \textit{Scikit-learn} library \citep{pedregosa2011scikit}, while neural models were trained on the \textit{Transformers} library \citep{wolf2020transformers} using the PyTorch backend running on Python 3.11.\footnote{Expect for N-shot GPT-4o baselines, which were run through OpenAI's API (\url{https://platform.openai.com/docs/api-reference/chat}), accessed on the 28th of July.}

\section{Experimental Results} 
\label{sec:results}

\subsection{Performance Evaluation}

\begin{table*}[h]
    \normalsize
    \centering
    \resizebox{\textwidth}{!}{
    \begin{tabular}{|l|p{2.3cm}|p{2.3cm}|p{2.3cm}|p{2.3cm}|p{2.3cm}|p{2.3cm}|}
    \hline
        \textbf{Model\textbackslash Dataset} & \multicolumn{2}{c|}{\textbf{OUM}} & \multicolumn{2}{c|}{\textbf{Wikitactics}} & \multicolumn{2}{c|}{\textbf{AFD}}\\
        \hline
        & \textbf{$r_{s}$}↑ & \textbf{MAE}↓ & \textbf{AUROC}↑ & \textbf{AUPR}↑ & \textbf{AUROC}↑ & \textbf{AUPR}↑\\
        \hline
        \multicolumn{7}{|c|}{\textbf{Simple Baselines}}\\
        \hline
        Average / Random  & 0 & 1.620 & 0.500 & 0.492 & 0.500 & 0.242 \\
        Bag-of-words & 0.233 (0.022) & 1.859 (0.006) & 0.679 (0.007) & 0.660 (0.003) & 0.883 (0.003) & 0.698 (0.008) \\
        \hline
        \multicolumn{7}{|c|}{\textbf{Neural Models}}\\
        \hline
        GloVe Embeddings & 0.393 (0.017) & 1.431 (0.011) & 0.590 (0.021) & 0.573 (0.021) & 0.826 (0.001) & 0.565 (0.003)\\
        Longformer (full) & 0.369 (0.023) & 1.454 (0.017)  & 0.541 (0.021) & 0.557 (0.020) & 0.845 (0.009) & 0.641 (0.015)\\
        Longformer (last layer) & 0.060 (0.007) & 1.654 (0.005)  & 0.502 (0.010) & 0.498 (0.012)  & 0.720 (0.002) & 0.412 (0.008) \\
        0-shot GPT-4o & 0.311 (0.012) & 1.901 (0.006)  & 0.644 (0.016) & 0.589 (0.011)  & 0.909 (0.003) & \textbf{0.855 (0.005)}\scriptsize{**} \\
        20-shot GPT-4o & 0.024 & 2.000  & 0.520 & 0.504  & 0.768 & 0.632 \\
        \hline
        \multicolumn{7}{|c|}{\textbf{(Standard) Feature-based Models}}\\
        \hline
        Politeness Markers & 0.262 (0.006) & 1.529 (0.004) & 0.688 (0.008) & 0.646 (0.009) & 0.757 (0.001) & 0.468 (0.007)\\
        Collaboration Markers & 0.251 (0.007) & 1.565 (0.003) & 0.705 (0.003) & 0.685 (0.002) & 0.735 (0.001)  & 0.422 (0.001) \\
        All Discrete Features & 0.318 (0.006) & 1.501 (0.005) &  0.706 (0.006) & 0.686 (0.003) & 0.760 (0.002)  &  0.454 (0.010) \\
        \hline
        \multicolumn{7}{|c|}{\textbf{LLM Feature-based Models}}\\
        \hline
        Dispute Tactics & 0.307 (0.009) & 1.516 (0.006) & 0.681 (0.012) & 0.683 (0.016) & 0.766 (0.003) & 0.463 (0.005)\\
        QoA & 0.011 (0.016) & 1.617 (0.002) & 0.657 (0.002) & 0.638 (0.005) & 0.526 (0.009) & 0.251 (0.012)\\
        Information Content & 0.105 (0.007) & 1.598 (0.001) & 0.553 (0.008) & 0.540 (0.013) & 0.598 (0.002) & 0.276 (0.004)\\
        Style and Tone &  0.269 (0.003) & 1.520 (0.002) & 0.612 (0.023) & 0.621 (0.015) & 0.907 (0.001) & 0.753 (0.003)\\
        All LLM-generated Features & 0.348 (0.009) & 1.468 (0.009) & 0.664 (0.009) & 0.652 (0.013) & 0.918 (0.000) & 0.772 (0.002)  \\
        All Discrete \& LLM-generated Features & 0.380 (0.003) & 1.443 (0.006) & \textbf{0.708 (0.004)}\scriptsize{**} & \textbf {0.690 (0.003)}\scriptsize{**} & \textbf{0.921 (0.002)}\scriptsize{*} & 0.791 (0.001) \\
            \hspace{0.2cm} \textit{$+$ participant-disaggregation} & \textbf{0.447 (0.006)}\scriptsize{*} & \textbf{1.421 (0.010)} & - & - & - & - \\
        \hline
        
    \end{tabular}
    }
    \caption{Predictive accuracy of the models considered. We use the paired t-test to compute the statistical difference between our best LLM feature-based model and the strongest neural baseline on each dataset. *p-value < 0.05 \& **p-value < 0.01.}
    \label{tab:performance_evaluation}
\end{table*}

As shown in Table \ref{tab:performance_evaluation}% shows the performance comparison of the baselines (average/random, bag-of-words, GloVe embeddings and two variants of Longformer) and the feature-based models (when taking distinct combinations of input features)
, the most performant baselines differ across datasets, suggesting the brittleness of neural models' performance. Meanwhile, the best LLM feature-based model is consistently the one that uses all (discrete and LLM-generated) features. It performs comparably (i.e., $p>0.05$; paired t-test) to the strongest baseline on OUM (GloVe Embeddings) and the strongest baseline on AFD (0-shot GPT-4o), and significantly outperforms all baselines on Wikitactics. In addition, if we disaggregate the features at the participant level, the LLM feature-based model significantly surpasses the strongest baseline (GloVe Embeddings) on the OUM dataset in terms of Spearman correlation ($p=0.02$; paired t-test). Interestingly, %if we compare the predictive power of different feature sets, we see that the optimal feature set also changes (i.e., ‘Dispute Tactics,’ ‘Collaboration Markers,’ and ‘Style and Tone’ are the most predictive feature sets for OUM, Wikitactics, and AFD, respectively). Likewise, the least predictive feature set varies depending on which dataset we analyse. In general, 
the ranking according to accuracy for different feature sets significantly changes for different datasets.

The results suggest three key insights: (1) LLM feature-based models built with the proposed framework can prove more accurate, or at least comparable, in comparison with both standard feature-based and neural models; (2) the framework may serve as a foundation and be extended by incorporating dataset-specific features like features disaggregated at the participant-level to enhance model performance further; (3) certain features can be more relevant than others depending on the application domains, and thus a framework incorporating a rich array of dataset-independent linguistic features, as proposed in this work, is valuable for dialogue constructiveness analysis. We next investigate whether the models have learned spuriously correlated shortcuts or robust decision rules%that may provide valuable insights into dialogue constructiveness analysis
.

\subsection{Performance Robustness Analysis}
\label{sec:shorcut_identification}

The OUM dataset is a suitable testbed for verifying shortcut learning due to its diversity in discussion topics (i.e., Brexit, vaccination, and veganism). We hypothesise that the performance of neural models would vary significantly when evaluated on different topics. To test this, we compare the performance of the best neural baselines, `Longformer (full)', `GloVe embeddings' and `0-shot GPT-4o', against our best LLM feature-based models across the different topics, using the same dataset and models in Table \ref{tab:performance_evaluation}, i.e., we simply split the dataset into three subsets representing different topics and calculate the metrics for each of the topics separately. As a result, neural models exhibit drastic performance degradation (mostly close to random) when assessed on individual topics (Table~\ref{tab:shortcut_evaluation})%, in spite of their overall competitive performance with respect to the feature-based models when evaluated on aggregate data, as we saw before in Table \ref{tab:performance_evaluation}
. 
%More concretely, the correlation scores for Longformer and GloVe models on all three topics are close to zero, indicating that these models' predictions are basically random when the data is disaggregated by topic. 
Thus, these neural models (GloVe Embeddings and Longformer) seem to indeed rely heavily on undesirable shortcuts. Upon a further investigation, we find that GloVe Embeddings and Longformer predict scores close to the mean of the score distribution of the input dialogue's topic, implying that these neural models have learned to identify which topic an input dialogue belongs to instead of robust prediction rules. This helps gain high correlation when topics are aggregated but null otherwise.

On the other end, our LLM feature-based models manifest robust performance across all topics, with significantly higher correlation scores and more stable MAE values. Noteworthy, on the Veganism topic, even if the neural models score random in terms of Spearman correlation, their MAE is lower than that of the LLM feature-based models. This can be attributed to the low standard deviation (1.335) within this topic, which allows neural models to achieve low error by simply predicting values close to the mean, albeit without accurately ranking the data points.% Hence, our feature-based models are much more robust for dialogue constructiveness prediction.

\begin{table*}[h]
\normalsize
\centering
\resizebox{\textwidth}{!}{
\begin{tabular}{|l|p{2.3cm}|p{2.3cm}|p{2.3cm}|p{2.3cm}|p{2.3cm}|p{2.3cm}|}
\hline
\textbf{Model\textbackslash Topic} & \multicolumn{2}{c|}{\textbf{Brexit}} & \multicolumn{2}{c|}{\textbf{Vaccination}} & \multicolumn{2}{c|}{\textbf{Veganism}}\\
\hline
& \textbf{$r_{s}$}↑ & \textbf{MAE}↓ & \textbf{$r_{s}$}↑ & \textbf{MAE}↓ & \textbf{$r_{s}$}↑ & \textbf{MAE}↓\\
\hline
\multicolumn{7}{|c|}{\textbf{Neural Models}}\\
\hline
GloVe & -0.036 (0.023) & 1.696 (0.007) & 0.043 (0.050) & 1.606 (0.017) & 0.107 (0.033) & \textbf{1.079 (0.009)}\scriptsize	{**} \\
Longformer (full) & 0.011 (0.043) & 1.756 (0.034) & 0.058 (0.016) & 1.623 (0.016) & -0.005 (0.038) & 1.084 (0.024)\\
0-shot GPT-4o & 0.438 (0.025) & 1.671 (0.029) & 0.054 (0.015) & 1.847 (0.028) & 0.244 (0.017) & 2.119 (0.022)\\
\hline
\multicolumn{7}{|c|}{\textbf{LLM Feature-based Models}}\\
\hline
All Discrete \& LLM-generated Features & 0.409 (0.055) & 1.428 (0.066) & 0.385 (0.015) & 1.444 (0.042) & 0.350 (0.050) & 1.449 (0.029)\\
\hspace{0.2cm} \textit{$+$ participant-disaggregation} & \textbf{0.445 (0.024)} & \textbf{1.405 (0.073)}\scriptsize	{**} & \textbf{0.461 (0.017)}\scriptsize	{**} & \textbf{1.381 (0.009)}\scriptsize	{**} & \textbf{0.433 (0.038)}\scriptsize	{**} & 1.411 (0.058)\\
\hline
\end{tabular}
}
\caption{Accuracy comparison between the most performant neural models and LLM feature-based models in Table \ref{tab:performance_evaluation} across three topics on OUM. We use the paired t-test to compute the statistical difference between our best LLM feature-based model and the strongest neural baseline on each dataset. *p-value < 0.05 \& **p-value < 0.01.}
\label{tab:shortcut_evaluation}
\end{table*}

\subsection{Explainable AI Analysis}
\label{sec:XAI}

% \subsection{Interpreting Feature-based Models}

%To interpret the predictive power of distinct linguistic features in predicting dialogue constructiveness, 
We interpret the decision-making of the best LLM feature-based models (without participant-disaggregation on the input features), by analysing the top-10 feature coefficients with the highest Feature Importance (FI), as quantified by their normalised FI (\%), computed via the permutation method\footnote{We use the `permutation\_importance' in https://scikit-learn.org/stable/modules/generated/sklearn.inspection.html. It assesses each feature's importance by shuffling the feature's values and measuring the model's performance change.} (Table \ref{tab:XAI_coeffs} in Appendix \ref{app:xai_analysis}).

For the OUM dataset, when predicting post-conversation open-mindedness (7-point scale), the feature with the highest importance (FI = 13.79\%) is the QoA, with a negative coefficient (-0.203). This negative sign is surprising, e.g., one might expect discussions that involve disagreement tactics like `DT - name calling/hostility' to receive very low QoA scores, leading to low open-mindedness. Indeed, data show that `DT - Name calling/hostility, $\bar{x}$' is slightly negatively correlated with both QoA ($r_s$ = -0.219) and `post-conversation open-mindedness' ($r_s$ = -0.135). % Namely, conversations without such impolite disagreement tactics should have slightly higher (or at least invariant) post-conversation open-mindedness scores. 
However, this contradicts the negative coefficient of QoA, which may be attributed to only 17.2\% of discussions using such impolite disagreement tactics, and thus, the negative coefficient depends on other nuanced details instead. Further analysis reveals that 93.7\% of conversations have a QoA score between 3 and 7.5%, i.e., very few discussions have extremely poor or good quality scores
. Scores 3-4 tend to be discussions where the participant focuses on asking questions, while scores $\geq 7$ tend to involve participants that argue with the chatbot or WoZ; this is supported by a weak negative correlation ($r_s$ = -0.306) between `DT - Asking questions, $\bar{x}$' and the QoA. This may be explained in two ways. First, information-seeking behaviour may leave more room for promoting post-conversation open-mindedness as the participant gains a better understanding of different viewpoints; it may also be that people who exhibit information-seeking behaviour tend to be more open-minded in the first place. This is supported by \citet{haran2013role}, which found that open-minded thinking positively correlates with information acquisition. Second, presenting strong arguments may signal a lower likelihood of considering opposing viewpoints or of increasing one's open-mindedness in believing people with opposing views have \textit{good reasons}, especially if the other side does not present highly convincing arguments. To confirm this, we inspect numerous instances with high QoA ($\geq 7$), where we extract two subsets with low ($\leq 2$) and high ($\geq 6$) open-mindedness scores, and observe contrasting patterns: (i) In the low open-mindedness subset, the WoZ/chatbot usually states some rationale for views opposing the participant’s belief but with weak or poor arguments and rarely challenges the participant, letting them to express more well-thought-out and convincing claims; (ii) In the high open-mindedness subset, the WoZ/chatbot offers longer, more diverse, and convincing arguments, with a stronger tendency to push back, and demonstrates a deeper understanding of the topic. All these suggest that the argument convincingness of the WoZ/chatbot for the participant’s opposing view seems to have a high impact on the post-conversation open-mindedness. Nonetheless, the coefficient of QoA is close to zero and should be interpreted with caution.

The second and third most influential features are the presence of hedge words (PM - Hedge words, $\bar{x}$; FI = 10.76\%) and the number of hedging terms (CM - \# Hedging terms, $\bar{x}$; FI = 10.37\%). Surprisingly, the former has a negative coefficient of -1.731, while the latter has a positive coefficient of 0.735. This discrepancy might stem from the difference in heuristics employed, supported by their moderate (Spearman) correlation of 0.547. For example, `PM - Hedge words' focuses on the presence of formal, objective, and subtly uncertain hedges % typical in academic writing 
terms (e.g., `estimate', `suggest', `likely'), while `CM - \# Hedging terms' counts both overlapping terms and additional conversational, personal, and explicitly uncertain hedges (e.g., `I think', `we hope', `good chance') that do not appear in the former. Further, the terms in `PM - Hedge words' focuses on implying moderate uncertainty (e.g., `seems', `typically', `mostly'), whereas `M - \# Hedging terms' also has additional terms (e.g., `not sure', `I guess') that directly express doubt. The effect of expressing %uncertainty through 
hedging language on predicting post-conversation open-mindedness is thus more nuanced % (depending on the types of hedging terms)
and warrants cautious interpretation. This conclusion is somehow similar to the work by \citet{vlasyan2018linguistic}, describing how nuanced the relationship between hedging language and politeness can be (e.g., hedges serve as both negative and positive politeness strategies in discourse).

%The degree of contextualisation (DT - Contextualisation, $\bar{x}$; FI = 10.32\%) has the fourth highest importance, with a negative coefficient of -2.193. One speculation is that providing excessive context alone might entrench participants in their existing beliefs, reducing post-conversation open-mindedness. 
Other features like negative and positive sentiments, have intuitively negative and positive coefficients% with respect to post-conversation open-mindedness
. Overall, many variables are relevant in predicting post-conversation open-mindedness, and interpreting them carefully could provide subtle, interesting insights.

We complete our XAI analysis on the Wikitactics and AFD datasets to Appendix \ref{app:xai_analysis}. Overall, analysing the most important features identified by LLM feature-based models can bring insights into key linguistic drivers of constructive dialogue. There is no clear evidence that our LLM feature-based models learn significant shortcuts, i.e., they tend to learn robust prediction rules. 

Interestingly, no single feature or feature set dominates across all datasets. This supports one of the concerns that motivated our work in the first place: Without a comprehensive list of linguistic features, like what we propose in our framework, it can be challenging and time-consuming to find relevant predictive signals if starting from scratch for a new dataset of dialogue constructiveness analysis. Noteworthy, few gradient-type ($\nabla$) features appear in the top-10 features. Still, it will be worth exploring them (and other statistics beyond average and gradient) in future work.

\section{Conclusions}

In research on dialogue constructiveness, the seemingly inevitable trade-off between cost, interpretability, and model performance has long been a problem. To break this curse, we propose a framework that provides a battery of dataset-independent linguistic features and can be leveraged to build LLM feature-based models for dialogue constructiveness assessment at a reasonably low cost. In modelling dialogue constructiveness across three datasets—Opening-up Minds, Wikitactics, and Articles for Deletion corpus—our results show convincing evidence that the LLM feature-based models %, which leverage linguistic features automatically generated by both GPT-4 and simple algorithmic rules, 
consistently outperform or performs as well as state-of-the-art neural models like GloVe, Longformer and N-shot GPT-4o variants, except for one single rare case in the AFD where our LLM feature-based model performed comparably with respect to the strongest baseline (0-shot GPT-4o). In addition, the LLM feature-based models tend to learn more robust decision rules than those learned by the neural models; the latter has a higher propensity on relying superficial shortcuts. Further, interpreting our LLM feature-based models can lead to both intuitive findings as well as several unexpected but illuminating discoveries regarding what linguistic factors influence the constructiveness of dialogues. The proposed framework thus serves as a valuable foundational toolkit for dialogue constructiveness research, helping researchers develop models with higher accuracy, robustness and interpretability in dialogue constructiveness assessment.

\section*{Limitations}

Our work is not without limitations. One is the automatic generation of features using LLMs, which require computational cost, despite being significantly more cost-effective than human annotations. This cost comparison is based on the following calculations: We have approximately 28K utterances across the three datasets. Let's assume the annotation cost is £10 per hour for hiring crowdsourced workers. Without considering the cost associated with any ethical committee approval, pilot studies and quality checks, the cost of annotating \textit{only} the `Dispute Tactics' feature set can easily exceed £3100, assuming it takes an average of 40 seconds to annotate one complete utterance. In contrast, the API querying cost associated with prompting the latest version of GPT-4, as of August 2024, for \textit{all} feature sets in our framework is below £500.  In addition to GPT-4, we also explored the option of prompting LLaMA-3.1-70B-Instruct and LLaMA-3.1-405B-Instruct for the feature extraction via HuggingFace API with a month of HuggingFace pro subscription, which only cost \$9, for annotating \textit{all} feature sets in about two weeks. This option was found to be moderately worse than using GPT-4 as the feature extractor. Nevertheless, we anticipate that the cost of prompting LLMs to generate linguistic features will become significantly lower in the near future, especially as open-source models become increasingly more powerful. Another limitation of our work lies in that we do not make a comparison between our LLM feature-based models and the State-of-The-Art (SOTA) model in the AFD dataset \citep{mayfield2019analyzing}. This is because here, we use a sample of 1000 data instances from the original data dataset, and thus, the comparison would be incommensurate. Still, some of the baselines (e.g., Longformer) are close re-implementations of the SOTA model (i.e., BERT), allowing us to demonstrate how competitive our approach is against SOTA, even though we were initially not focusing on beating SOTA.

There are several promising research avenues about our framework that have not been covered in this work, which we encourage future efforts. For instance, future research should be encouraged to investigate the inclusion of more dataset-specific features that can enhance model performance further (e.g., features at the participant level, individual personalities and demographic attributes), which may provide predictive power and additional insights into the dynamics of dialogue constructiveness. Besides, leveraging other statistics beyond the average and gradient statistics (used in this work) may uncover additional predictive signals. Another promising direction is the leveraging of our framework (with certain adaptions) to other domains and dialogue types beyond dialogue constructiveness analysis.

\section*{Acknowledgments}

This work has been supported by the Open Philanthropy Long-term Future Fund and EPSRC grant no. EP/T023414/1: Opening Up Minds. This research project has also benefited from the Microsoft Accelerate Foundation Models Research (AFMR) grant program.

\bibliography{refs}

\appendix

\section{Appendix}

\subsection{Detailed Feature Descriptions}
\label{app:feature_descriptions}

Here, we present detailed definitions and/or examples of all individual linguistic features used in our experiments. Table \ref{tab:politeness-markers} and \ref{tab:collaboration-markers} provide the definition of features in `politeness markers' and `collaboration markers' feature sets, respectively. Table \ref{tab:dt_rt} and \ref{tab:dt_ct} define distinct rebuttal tactics and collaborative tactics in the `dispute tactics' feature set. Table \ref{tab:info_dens_and_style_tones} describes both `information content' and `style and tone' features with examples.

\begin{table*}[h!]
    \centering
    \footnotesize
    \begin{tabular}{|p{4cm}|p{11cm}|}
        \hline
        \textbf{Feature} & \textbf{Description} \\
        \hline
        Please & The presence of the word `please' in the sentence. \\
        \hline
        Start with `Please' & The sentence starts with the word `please'. \\
        \hline
        Has subject hedge & Any subject in the sentence depends on a hedge word. \\
        \hline
        Use of `by the way' & The phrase `by the way' is used in the sentence. \\
        \hline
        Hedge words & Any word in the sentence is a hedge word. \\
        \hline
        Assert factuality & Words that assert factuality, like `in fact,' `actually,' or `really'. \\
        \hline
        Start with deference & The sentence starts with deferential words like `great,' `good,' or `nice'. \\
        \hline
        Gratitude & Expressions of gratitude, like `thank' or `thanks'. \\
        \hline
        Apologising & Apologetic expressions like `sorry,' or `I apologize'. \\
        \hline
        1st person plural & First-person plural pronouns. \\
        \hline
        1st person pronouns & First-person singular pronouns. \\
        \hline
        Start with 1st person & The sentence starts with a first-person singular pronoun. \\
        \hline
        2nd person pronouns & Second-person pronouns like. \\
        \hline
        Start with 2nd person & The sentence starts with a second-person pronoun. \\
        \hline
        Start with greeting & The sentence starts with a greeting word. \\
        \hline
        Starts with question & The sentence starts with a question word like `what,' `why,' `who,' or `how'. \\
        \hline
        Starts with conjunction & The sentence starts with a conjunction or transition word like `so,' `then,' `and,' `but,' or `or'. \\
        \hline
        Positive sentiment words & The presence of positive sentiment words. \\
        \hline
        Negative sentiment words & The presence of negative sentiment words. \\
        \hline
        Subjunctive words & The use of subjunctive mood words like `could' or `would' when preceded by `you'. \\
        \hline
        Indicative words & The use of indicative mood words like `can' or `will' when preceded by `you'. \\
        \hline
    \end{tabular}
    \caption{Descriptions of politeness markers. For some features, we provide a few examples for clarification. For the whole list of examples, please visit the original paper \citep{zhang2018conversations}.}
    \label{tab:politeness-markers}
\end{table*}

\begin{table*}[h!]
    \centering
    \footnotesize
    \begin{tabular}{|p{4cm}|p{11cm}|}
        \hline
        \textbf{Feature} & \textbf{Description} \\
        \hline
        \# words & The number of words in an utterance.\\
        \hline
        \# me pronoun & The number of first-person singular pronouns.\\
        \hline
        \# we pronoun & The number of first-person plural pronouns.\\
        \hline
        \# you pronoun & The number of second-person pronouns.\\
        \hline
        \# 3rd person pronouns & The number of third-person pronouns.\\
        \hline
        \# Geography terms & The number of geography-related terms. \\
        \hline
        \# Meta terms & The number of meta-discourse terms. Meta terms are associated with the functionalities, actions, and elements within the StreetCrowd environment. \\
        \hline
        \# Certainty terms & The number of words expressing certainty. \\
        \hline
        \# Hedging terms & The number of hedging terms (words that indicate uncertainty). \\
        \hline
        \# New content words & The number of new content words introduced by a user. New content words are content words (significant words such as nouns, verbs, adjectives) that have not been seen in any previous messages. These exclude common stopwords and are identified based on part-of-speech tags. \\
        \hline
        \# New content words * \# Certainty terms & The number of new content words introduced in an utterance that are also accompanied by certainty terms. The feature is calculated as the product of the number of new content words (\# New content words) and the count of certainty terms (\# Certainty terms) in the message. \\
        \hline
        \# New content words * \# Hedging terms & The number of new content words introduced in an utterance that are also accompanied by hedging terms. It is computed as the product of the number of new content words (\# New content words) and the count of hedging terms (\# Hedging terms) in the message. \\
        \hline
    \end{tabular}
    \caption{Descriptions of collaboration markers. For further details, please visit the original paper \citep{niculae2016conversational}.}
    \label{tab:collaboration-markers}
\end{table*}

\begin{table*}[]
    \centering
    \footnotesize
    \begin{tabular}{|l|p{11.5cm}|}
    \hline
        \textbf{Feature} & \textbf{Description} \\\hline
        Name calling/hostility & Direct insults, or use of an equally hostile tone or language.\\\hline
        Ad hominem/ad argument & An attack to the person, often used to attempt to discredit an opponent (eg.\ ``I know better than you; you do not have a physics degree''). \citet{de2022disagree} extended this class to include ``ad argument'', where someone insults another's argument (eg.\ ``that is ridiculous'') without responding to its content.\\\hline
        Attempted derailing/off-topic & This category was added to address comments which are unrelated to the current line of discussion and fail to further the argument, while still being argumentative. It was assigned a lower level as it can be detrimental to the argument, by taking focus away from the main topic. \\\hline
        Policing the discussion & \citet{graham2008disagree} referred to this as “responding to tone” with the description “responses to the writing, rather than the writer. The lowest form of these is to disagree with the author's tone”. By expanding the definition to “policing the discussion”, the idea is to include people saying “You’ve said that before”, telling people to “calm down”, correcting spelling errors, or citing discussion policy (ie. “no personal attacks”). It ignores the argument’s content. \\\hline
        Stating your stance & \citet{graham2008disagree} referred to this as ``contradiction'', with the description ``to state the opposing case, with little or no supporting evidence''. \citet{de2022disagree} renamed this class to include editors joining the conversation and just voicing their stance or their agreement with another editor.\\\hline
        Repeated argument & Level added to describe re-stating an argument used before, potentially using different words. This level was chosen as, similar to Level `Stating your stance': Stating your stance, it does engage with the argument but does not further the discussion. \\\hline
        Counterargument & Described as ``contradiction plus new reasoning and/or evidence'', which does not directly address the opponent's argument.\\\hline
        Refutation & Directly responding to the argument and explaining why it is mistaken, using new evidence or reasoning.\\\hline
        Refuting the central point & \citet{graham2008disagree} notes that ``Truly refuting something requires one to refute its central point, or at least one of them.'' Unfortunately, the central point can be quite subjective and difficult to recognise for non-experts, and may change throughout a conversation. As such, we use this category as a prime example of a ``good'' refutation, which is of course still subjective and may be rolled up into `Refutation'. \\
        \hline
    \end{tabular}
    \caption{Rebuttal tactics descriptions. Adapted from \citep{de2022disagree}.}
    \label{tab:dt_rt}
\end{table*}

\begin{table*}[]
    \centering
    \footnotesize
    \begin{tabular}{|l|p{12cm}|}
        \hline
        \textbf{Feature} & \textbf{Description}\\
        \hline
        Bailing out&An indication that an individual is giving up on a conversation and will no longer engage.\\\hline
        Contextualisation& Usually in the first utterance, an individual ``sets the stage" by describing what aspect they are challenging. This does not directly disagree with anyone, and is therefore a non-disagreement move.\\\hline
        Asking questions&Seeking to understand another individual's opinion better. This does not include rhetorical questions, which are generally disagreement moves.\\\hline
        Providing clarification&Answering questions or providing information which seeks to create understanding, rather than only furthering a point.\\\hline
        Suggesting a compromise&An attempt to find a midway between one's own point and the opposer's. \\\hline
        Coordinating & Take the example of Wikipedia Talk pages, which are primarily used to for goal-oriented discussions, to coordinate edits to a page. As part of disagreement threads, there is often also some discussion of these edits. This can signal that a compromise has been found.\\\hline
        Conceding / recanting&An explicit admission that an interlocutor is willing to relinquish their point.\\\hline
        I don't know&Admitting that one is uncertain. This signals that an editor is receptive to the idea that there are unknowns which may impact their argument.\\\hline
        Other&For utterances not covered by any other class, for instance social niceties.\\
        \hline
    \end{tabular}
    \caption{Coordination tactics descriptions. Adapted from \citep{de2022disagree}.}
    \label{tab:dt_ct}
\end{table*}

\begin{table*}[]
    \centering
    \footnotesize
    \begin{tabular}{|l|p{3cm}|p{9.5cm}|}
    \hline
        \textbf{Feature} & \textbf{Description} & \textbf{Levels \& Example Sentences} \\\hline
        
        P-DENSITY & Measures propositional density, the ratio of propositions to the number of words in a text. & 
        \textbf{Low}: In addition to what I suggested earlier about testing for the non-existence of a third file, we could also verify that the contents of the sync database files are not nonsense. \newline
        \textbf{High}: I tried patching this in locally and it doesn’t compile. \\\hline
        
        C-DENSITY & Measures content density, the ratio of open-class to closed-class words. & 
        \textbf{Low}: Slight reordering: please put system modules first, then a blank line, then local ones (PRESUBMIT). \newline
        \textbf{High}: Please check that given user id is child user, not currently active user is child. \\\hline
        
        FORMALITY & Measures formality using a binary (formal vs. informal) scale. & 
        \textbf{Low}: But yeah, I’m just being an API astronaut*; I think that what I wrote up there is neat, but after sleeping, don’t worry about it; it’s too much work to go and rewrite stuff. \newline
        \textbf{High}: Moving this elsewhere would also keep this module focused on handling the content settings / heuristics for banners, which is what it was originally intended for. \\\hline
        
        POLITENESS & Measures politeness using annotated corpus models. & 
        \textbf{Low}: You don’t actually manage the deopt table’s VirtualMemory, so you shouldn’t act like you do. \newline
        \textbf{High}: Thanks for writing this test, getting there, but I think you could do this in a more principled way. \\\hline
        
        SENTIMENT & Measures the sentiment of a comment (negative, neutral, positive). & 
        \textbf{Negative}: That’s not good use of inheritance. \newline
        \textbf{Neutral}: Are we planning on making use of this other places? \newline
        \textbf{Positive}: It looks slightly magical. \\\hline
        
        UNCERTAINTY & Measures types of uncertainty exhibited in comments. & 
        \textbf{Epistemic}: This seems a bit fragile. \newline
        \textbf{Doxastic}: I assume we added this notification purely for testing purposes? \newline
        \textbf{Investigative}: Did you check whether it was needed? \newline
        \textbf{Conditional}: Another possible option, if it does not cause user confusion, would be to automatically select those projects in the Files view when the dialog closes. \\
        \hline
    \end{tabular}
    \caption{Descriptions about `information content' and `style and tone'. The definitions and examples are adapted/sourced from \citep{meyers2018dataset}.}
    \label{tab:info_dens_and_style_tones}
\end{table*}

\subsection{Prompts to Generate Features}
\label{app:prompts_generating_feats}

This appendix contains the prompt templates used to generate all LLM-generated features included in our analysis. Table \ref{tab:prompt_dispute_tactics} delineates the prompt used for generating the `dispute tactics' feature set. Table \ref{tab:prompt_quality_of_arguments} presents one example prompt used for generating the `quality of arguments' feature set. Table \ref{tab:prompt_clrf} shows a prompt for generating `information content' and `style and tone' feature sets.

\subsection{Annotation Quality of LLaMA-3.1 LLMs}
\label{ap:llama3dot1_annot_quality}

In Table \ref{tab:annotation_quality_comparison}, we compare the annotation quality between GPT-4, LLaMA-3.1-70B-Instruct, and LLaMA-3.1-405B-Instruct, evaluated for the whole three datasets. Overall, GPT-4 and LLaMA-3.1-405B-Instruct vastly outperform LLaMA-3.1-70B-Instruct. GPT-4 moderately surpasses LLaMA-3.1-405B-Instruct in terms of the number of times it outperforms other models for each tuple of <dataset, feature set, metric>, with GPT-4 topping 13 times, while LLaMA-3.1-405B-Instruct tops 9 times. The dominance of GPT-4 is especially pronounced in the AFD dataset\footnote{We further investigated this by inspecting the annotations of LLaMA-3.1-405B-Instruct in AFD and found that the annotation error is considerably higher than GPT-4 in this dataset. For instance, the former sometimes annotates both “low politeness” and “high politeness” in the “Style and tone” feature set as 0 when exactly one feature should be annotated as 1.}.

Notably, we also found that the \textit{instructability} of these two LLaMA-3.1 models is worse than that of GPT-4 during the experiment, especially for LLaMA-3.1-70B-Instruct. For example, both models sometimes required an additional sentence to produce the desired annotations. Otherwise, the models might repeat the question or include the prompt by specifying more rules for the annotation of features. We thus, for some cases, added an additional sentence such as “\textbackslash{}nAnnotations:” or “\textbackslash{}nChain-of-Thought reasoning step:” to obtain the desired outputs\footnote{This makes it slightly unfair for GPT-4, and thus should be taken into account when interpreting Table \ref{tab:annotation_quality_comparison}.}. Another interesting observation was that LLaMA-3.1-70B does not yield a number when annotating QoA in less than 2\% of cases in OUM, and `dispute tactics' feature set in about 1\% of cases in AFD. Such small amount of missing data was imputed with the mean. This is only a limitation of the smaller model, as it no longer presents in the larger 405B version. All things considered, we use GPT-4 to annotate the features for our main analyses.

\begin{table*}[h]
    \normalsize
    \centering
    \resizebox{\textwidth}{!}{
    \begin{tabular}{|l|p{2.3cm}|p{2.3cm}|p{2.3cm}|p{2.3cm}|p{2.3cm}|p{2.3cm}|}
    \hline
        \textbf{Feature Set\textbackslash Dataset} & \multicolumn{2}{c|}{\textbf{OUM}} & \multicolumn{2}{c|}{\textbf{Wikitactics}} & \multicolumn{2}{c|}{\textbf{AFD}}\\
        \hline
        & \textbf{$r_{s}$}↑ & \textbf{MAE}↓ & \textbf{AUROC}↑ & \textbf{AUPR}↑ & \textbf{AUROC}↑ & \textbf{AUPR}↑\\
        \hline
        \multicolumn{7}{|c|}{\textbf{GPT-4}}\\
        \hline
        Dispute Tactics & \textbf{0.307 (0.009)} & \textbf{1.516 (0.006)} & 0.681 (0.012) & \textbf{0.683 (0.016)} & \textbf{0.766 (0.003)} & \textbf{0.463 (0.005)}\\
        QoA & 0.011 (0.016) & \textbf{1.617 (0.002)} & \textbf{0.657 (0.002)} & \textbf{0.638 (0.005)} & 0.526 (0.009) & 0.251 (0.012)\\
        Information Content & \textbf{0.105 (0.007)} & 1.598 (0.001) & 0.553 (0.008) & 0.540 (0.013) & \textbf{0.598 (0.002)} & \textbf{0.276 (0.004)}\\
        Style and Tone &  0.269 (0.003) & 1.520 (0.002) & 0.612 (0.023) & 0.621 (0.015) & \textbf{0.907 (0.001)} & \textbf{0.753 (0.003)}\\
        \hline
        \multicolumn{7}{|c|}{\textbf{LLaMA-3.1-70B-Instruct}}\\
        \hline
        Dispute Tactics  & 0.280 (0.006) & 1.553 (0.020) & 0.642 (0.011) & 0.623 (0.011) & 0.513 (0.013) & 0.262 (0.012)\\
        QoA  & -0.090 (0.021) & 1.621 (0.002) & 0.625 (0.001) & 0.610 (0.002) & \textbf{0.659 (0.002)} & \textbf{0.386 (0.001)}\\
        Information Content & 0.057 (0.006) & 1.609 (0.003) & 0.379 (0.008) &  0.434 (0.006) & 0.526 (0.007) &  0.265 (0.007)\\
        Style and Tone  & 0.282 (0.014) & 1.537 (0.007) & 0.626 (0.013) & 0.658 (0.005) & 0.510 (0.006) & 0.246 (0.005)\\
        \hline
        \multicolumn{7}{|c|}{\textbf{LLaMA-3.1-405B-Instruct}}\\
        \hline 
        Dispute Tactics  & 0.247 (0.024) & 1.570 (0.018) & \textbf{0.695 (0.011)} & 0.670 (0.014) & 0.515 (0.016) & 0.244 (0.012)\\
        QoA  & \textbf{-0.106 (0.031)} & 1.622 (0.003) & 0.593 (0.003) & 0.569 (0.004) & 0.646 (0.003) & 0.333 (0.004) \\
        Information Content & 0.025 (0.013) & \textbf{1.544 (0.011)} & \textbf{0.576 (0.005)} & \textbf{0.546 (0.021)} & 0.440 (0.007) & 0.211 (0.004) \\
        Style and Tone  & \textbf{0.376 (0.011)} & \textbf{1.458 (0.006)} & \textbf{0.669 (0.011)} & \textbf{0.660 (0.015)} & 0.498 (0.019) & 0.236 (0.009) \\
        \hline
        
    \end{tabular}
    }
    \caption{Comparison of annotation quality between GPT-4, LLaMA-3.1-70B-Instruct and LLaMA-3.1-405B-Instruct. Each tuple of <dataset, feature set, metric> is compared across the three models and the most performance model is bolded.}
    \label{tab:annotation_quality_comparison}
\end{table*}

\subsection{Error Analysis on the LLM's annotations}
\label{ap:error_analysis}

Inspecting the annotated features compared with human annotations, we find that although the accuracy of LLM-derived features (excluding QoA) ranged between 86-100\%, most were >95\%, with a small subset of features presenting moderately lower accuracy (e.g., ‘DT - Stating your stance’, ‘Formality’). For example, GPT-4 usually fails when the speaker’s expression/intention is vague (e.g., an individual ambiguously states their stance with a comment of “2XL>\textbackslash{}n*VOTE” in support of another editor named “2XL”) or a binary feature that has a vague boundary (e.g., ‘FORMALITY’). For QoA, disagreements between human and LLM judgement often happen when comparing a pair of articles with similar QoA scores (difference <=1), which makes sense since it would be hard to judge which article deserves a higher score of QoA when both have similar levels of quality of arguments.

\begin{table*}[]
    \centering
    \scriptsize
    \begin{tabular}{|p{15.5cm}|}
    \hline
Below are two sets of linguistic features for disagreement levels and non-disagreement labels.\textbackslash{}n

Disagreement Levels:\textbackslash{}n

Level 0: Name calling/hostility - Direct insults, or use of an equally hostile tone or language.\textbackslash{}n

Level 1: Ad hominem/ad argument - An attack to the person, often used to attempt to discredit an opponent (e.g., "I know better than you; you do not have a physics degree"). This category was extended to include "ad argument", where someone insults another's argument (e.g., "that is ridiculous") without responding to its content.\textbackslash{}n

Level 2: Attempted derailing/off-topic - This category was added to address comments which are unrelated to the current line of discussion and fail to further the argument, while still being argumentative. It was assigned a lower level as it can be detrimental to the argument, by taking focus away from the main topic.\textbackslash{}n

Level 3: Policing the discussion - Graham (2008) referred to this as "responding to tone" with the description "responses to the writing, rather than the writer. The lowest form of these is to disagree with the author’s tone". By expanding the definition to "policing the discussion", the idea is to include people saying "You’ve said that before", telling people to "calm down", correcting spelling errors, or citing discussion policy (i.e. "no personal attacks"). It ignores the argument’s content.\textbackslash{}n

Level 4: Stating your stance - Graham (2008) referred to this as "contradiction", with the description "to state the opposing case, with little or no supporting evidence".\textbackslash{}n

Level 4: Repeated argument - Level added to describe re-stating an argument used before, potentially using different words. This level was chosen as, similar to `Level 4: Stating your stance', it does engage with the argument but does not further the discussion.\textbackslash{}n

Level 5: Counterargument - Described as "contradiction plus new reasoning and/or evidence", which does not directly address the opponent’s argument.\textbackslash{}n

Level 6: Refutation - Directly responding to the argument and explaining why it is mistaken, using new evidence or reasoning.\textbackslash{}n

Level 7: Refuting the central point - Graham (2008) notes that "Truly refuting something requires one to refute its central point, or at least one of them." Unfortunately, the central point can be quite subjective and difficult to recognize for non-experts, and may change throughout a conversation. As such, we use this category as a prime example of a “good” refutation, which is of course still subjective and may be rolled up into Level 6.\textbackslash{}n

Non-Disagreement Labels:

Label A: Bailing out - An indication that a person is giving up on a conversation and will no longer engage.\textbackslash{}n

Label B: Contextualisation - Usually in the first utterance, a person "sets the stage" by describing which aspect of the discussion they are challenging. This does not directly disagree with anyone, and is therefore a non-disagreement move.\textbackslash{}n

Label C: Asking questions - Seeking to understand another person’s opinion better. This does not include rhetorical questions, which are generally disagreement moves.\textbackslash{}n

Label D: Providing clarification - Answering questions or providing information which seeks to create understanding, rather than only furthering a point.\textbackslash{}n

Label E: Suggesting a compromise - An attempt to find a midway between one’s own point and the opposer’s.\textbackslash{}n

Label F: Coordinating - Take the example of Wikipedia Talk pages, which are primarily used to for goal-oriented discussions, to coordinate edits to a page. As part of disagreement threads, there is often also some discussion of these edits. This can signal that a compromise has been found.\textbackslash{}n

Label G: Conceding / recanting - An explicit admission that an interlocutor is willing to relinquish their point.\textbackslash{}n

Label H: I don't know - Admitting that one is uncertain. This signals that an person is receptive to the idea that there are unknowns which may impact their argument.\textbackslash{}n

Label I: Other - For utterances not covered by any other class, for instance, social niceties.\textbackslash{}n

Given a conversation history (which can be empty if the new utterance is the first utterance in the conversation), please analyse a new utterance from an individual (identified by a unique user\_id) in a conversation discussing with others about a potentially controversial edit from Wikipedia Talk pages:\textbackslash{}n

*CONVERSATION HISTORY*: "\texttt{\{CONVERSATION\_HISTORY\}}"\textbackslash{}n

*NEW UTTERANCE*: "\texttt{\{UTTERANCE\}}"\textbackslash{}n

Noteworthy, the conversation history is provided so you can simply understand the utterances delivered before the new utterance so as to help you better annotate the new utterance.\textbackslash{}n

Thus, please do not annotate the entire conversation but annotate only the new utterance by determining its appropriate disagreement level and/or non-disagreement label(s). Please provide the final answer directly with no reasoning steps.\textbackslash{}n

If the new utterance fits multiple categories, list all that apply. Ensure that your final answer clearly identifies whether each of the disagreement levels (0-7) and non-disagreement labels (A-I) are applicable (1) or not applicable (0) to the new utterance.\textbackslash{}n

For clarity, your response should succinctly be presented in a structured list format, indicating the presence (1) or absence (0) of each level and label as follows:

- Level 0: [1/0]

- Level 1: [1/0]

- Level 2: [1/0]

......

- Label G: [1/0]

- Label H: [1/0]

- Label I: [1/0]
    \\
    \hline
    \end{tabular}
    \caption{The prompt (to GPT-4) used for generating `dispute tactics' for the wikitactics dataset, given a new utterance alongside the conversation history.}
    \label{tab:prompt_dispute_tactics}
\end{table*}

\begin{table*}[]
    \centering
    \small
    \begin{tabular}{|p{15.5cm}|}
    \hline
    \texttt{\{INPUT\_DIALOGUE\}\textbackslash{}n\textbackslash{}n} \\
    The texts above show a dialogue with respect to a potentially controversial edit between two or more individuals from Wikipedia Talk pages. All individuals present arguments on why they think the edit is or isn't reasonable, possibly incorporating various rhetorical strategies, factual accuracy, relevance, coherence, among others. Now, please use chain-of-thought reasoning to evaluate the *average quality of arguments* (i.e., the discussion quality).
    After the Chain-of-Thoughts reasoning steps, you should assign an overall quality score to the entire discussion on a scale from 1 to 10, where 1 is of poor quality and 10 is of extremely good quality. Conclude your evaluation with the statement: 'Thus, the quality score of the discussion is: X', where X is the numeric score (real number) you've determined.\\
    \hline
    \end{tabular}
    \caption{An example prompt (to GPT-4) used for generating the feature `quality of arguments' for the wikitactics dataset, given an input dialogue. The prompts used for the OUM and AFD datasets are identical but changing the first sentence in the prompt to better contextualise where the input dialogue comes from.}
    \label{tab:prompt_quality_of_arguments}
\end{table*}

\begin{table*}[]
    \centering
    \scriptsize
    \begin{tabular}{|p{15.5cm}|}
    \hline
UTTERANCE: "\texttt{\{UTTERANCE\}}"\textbackslash{}n

Given the UTTERANCE above, please analyse it and extract the following linguistic features:\textbackslash{}n

Propositional Density:

- Propositional density is the ratio of propositions (i.e., distinct assertions or ideas) to the total number of words in the comment. It measures information content.

- Low Propositional Density Example: "In addition to what I suggested earlier about testing for the non-existence of a third file, we could also verify that the contents of the sync database files are not nonsense."

- High Propositional Density Example: "I tried patching this in locally and it doesn't compile."\textbackslash{}n

Content Density:

- Content density is the ratio of open-class words (e.g., nouns, verbs, adjectives, adverbs) to closed-class words (e.g., pronouns, prepositions, conjunctions). It is another measure of information content.

- Low Content Density Example: "Slight reordering: please put system modules first, then a blank line, then local ones (PRESUBMIT)."

- High Content Density Example: "Please check that given user id is child user, not currently active user is child."\textbackslash{}n

Formality:

- Formality assesses whether the language used in the comment is formal or informal.

- Low Formality Example: "But yeah, I'm just being an API astronaut*; I think that what I wrote up there is neat, but after sleeping, don't worry about it; it's too much work to go and rewrite stuff."

- High Formality Example: "Moving this elsewhere would also keep this module focused on handling the content settings / heuristics for banners, which is what it was originally intended for."\textbackslash{}n

Politeness:

- Politeness measures the degree of respect and courtesy expressed in the comment.

- Low Politeness Example: "You don't actually manage the deopt table's VirtualMemory, so you shouldn't act like you do."

- High Politeness Example: "Thanks for writing this test, getting there, but I think you could do this in a more principled way."\textbackslash{}n

Sentiment:

- Sentiment analysis determines the overall emotional tone of the comment.

- Negative Sentiment Example: "That's not good use of inheritance."

- Neutral Sentiment Example: "Are we planning on making use of this other places?"

- Positive Sentiment Example: "It looks slightly magical."\textbackslash{}n

Uncertainty Type:

- Uncertainty type identifies the specific type of uncertainty expressed in the comment, if any.

- Epistemic Uncertainty Example: "This seems a bit fragile."

- Doxastic Uncertainty Example: "I assume we added this notification purely for testing purposes?"

- Investigative Uncertainty Example: "Did you check whether it was needed?"

- Conditional Uncertainty Example: "Another possible option, if it does not cause user confusion, would be to automatically select those projects in the Files view when the dialogue closes."\textbackslash{}n

Please provide the final answer concisely and directly with no reasoning steps. For each feature, use 1 to indicate the presence and 0 to indicate the absence of the specific feature or feature option, as follows:

- Low Frazier Score: [0/1]

- High Frazier Score: [0/1]

- Low Yngve Score: [0/1]

- High Yngve Score: [0/1]

- Low Propositional Density: [0/1]

- High Propositional Density: [0/1]

- Low Content Density: [0/1]

- High Content Density: [0/1]

- Low Formality: [0/1]

- High Formality: [0/1]

- Low Politeness: [0/1]

- High Politeness: [0/1]

- Negative Sentiment: [0/1]

- Neutral Sentiment: [0/1]

- Positive Sentiment: [0/1]

- Epistemic Uncertainty: [0/1]

- Doxastic Uncertainty: [0/1]

- Investigative Uncertainty: [0/1]

- Conditional Uncertainty: [0/1]

- No Uncertainty: [0/1]\textbackslash{}n

If multiple Uncertainty Types are detected, mark all that apply as 1. If no Uncertainty is detected, mark No Uncertainty as 1 and the rest as 0.
    \\
    \hline
    \end{tabular}
    \caption{The prompt used for generating `information content' and `style and tone' feature sets, given an input utterance.}
    \label{tab:prompt_clrf}
\end{table*}

\subsection{Scaling the Number of Shots on GPT-4o}
\label{ap:gpt4o_scaling_analysis}

Here we analyse the effect of increasing the number of shots of the N-shot GPT-4o baselines, with N $\in \{0, 1, 2, 5, 10, 20\}$. The result is shown in Table \ref{tab:Nshot_GPT4o_performance_comparison} and the prompts are in Table \ref{tab:gpt4o_baseline_prompt}. Surprisingly, we observe a reasonably steady decline in baseline performance from 0 to 20-shot scenarios. This could be due to two intertwined hypotheses:

\begin{itemize}
    \item It might be easier to "\textit{look at the contextual information and predict the score of the test instance}" than "\textit{look at both contextual information and few provided examples to predict the score}" since GPT-4o may get distracted by the few shot (lengthy) examples and consequently pay less attention to the contextual information that might be critical to gain predictive power.
    \item There might be a limitation in GPT-4o’s attention mechanism, possibly using a fixed-size sliding window to mitigate the quadratic complexity in self-attention computation, which may restrict performance for tasks of long sequences but scale up the context size.
\end{itemize}

Given GPT-4o's proprietary nature, it would be insightful if more rigorous future work analysed these two hypotheses. Overall, we can at least conclude that increasing the number of shots above 20 will likely prove futile, and there is thus no need to scale the number of shots further to avoid being wasteful. It is also worth noting that the 20-shot prompted GPT-4o baseline is nearly as costly as our LLM feature-based models, and thus,  one should opt for the latter for pragmatic reasons.

On the other hand, if we compare the performance of the N-shot variants of GPT-4o with our best LLM feature-based models (Table \ref{tab:performance_evaluation}), we observe that the latter significantly outperform N-shot GPT-4o on all setups (zero/few-shots) in the OUM and Wikitactics datasets, by a notable margin. The sole exception is AFD, where our best LLM feature-based model surpasses all GPT-4o N-shot variants in AUROC, but scores moderately lower (though remains competitive) in AUPR in some cases. The exception in AFD might be attributed to two reasons: (1) Features like positive and negative sentiments provide high predictive power for AFD as shown in our XAI analysis (section \ref{sec:XAI}). Since LLMs usually excelled at sentiment analysis benchmarks, it would be unsurprising if GPT-4o were leveraging such information; (2) Potential data contamination, with GPT-4o’s training possibly including Wikipedia deletion debates, a common data source for LLM pre-training.

\begin{table*}[h]
    \normalsize
    \centering
    \resizebox{\textwidth}{!}{
    \begin{tabular}{|l|p{2.3cm}|p{2.3cm}|p{2.3cm}|p{2.3cm}|p{2.3cm}|p{2.3cm}|}
    \hline
        \textbf{Model\textbackslash Dataset} & \multicolumn{2}{c|}{\textbf{OUM}} & \multicolumn{2}{c|}{\textbf{Wikitactics}} & \multicolumn{2}{c|}{\textbf{AFD}}\\
        \hline
        & \textbf{$r_{s}$}↑ & \textbf{MAE}↓ & \textbf{AUROC}↑ & \textbf{AUPR}↑ & \textbf{AUROC}↑ & \textbf{AUPR}↑\\
        \hline
        0-shot & \textbf{0.311 (0.012)} & \textbf{1.901 (0.006) } & \textbf{0.644 (0.016)} & \textbf{0.589 (0.011)} & \textbf{0.909 (0.003)} & \textbf{0.855 (0.005)}\\
        1-shot & 0.114 & 1.996 & 0.548 & 0.528 & 0.875 & 0.743 \\
        2-shot & 0.055 & 1.942 & 0.578 & 0.548 & 0.895 & 0.791 \\
        5-shot & 0.000 & 2.055 & 0.508 & 0.506 & 0.907 & 0.835 \\
        10-shot & 0.001 & 2.098 & 0.499 & 0.506 & 0.889 & 0.805 \\
        20-shot & 0.024 & 2.000 & 0.520 & 0.504 & 0.768 & 0.632 \\
        \hline
        
    \end{tabular}
    }
    \caption{Performance of N-shot GPT-4o baselines. We only run 3 seeds for zero-shot setting since it is the best N-shot variant and to avoid being wasteful with compute. The used prompts for GPT-4o can be found in Table \ref{tab:gpt4o_baseline_prompt}.}
    \label{tab:Nshot_GPT4o_performance_comparison}
\end{table*}

\begin{table*}[]
    \centering
    \small
    \begin{tabular}{|p{15.5cm}|}
    \hline
    \multicolumn{1}{|c|}{\textbf{OUM}} \\
    \hline
    Predict the post-conversation open-mindedness score based on the given conversation. This score reflects a participant's self-reported level of open-mindedness towards opposing views on a controversial topic (veganism, Brexit, or vaccination) after discussing it with another entity (human or disguised chatbot).\textbackslash{}n   
    
    More specifically, the score is based on a 7-point Likert scale, where participants rate whether they believe people with opposing views to theirs have good reasons.\textbackslash{}n
    
    Key points:
    
    1. Score range: 1 to 7
    
    2. Higher scores indicate greater open-mindedness
    
    3. Scores were self-reported after the conversation
    
    4. Topics: veganism, Brexit, or vaccination\textbackslash{}n
    
    Important: Your response should consist of a single number, with no additional text.\textbackslash{}n
    
    CONVERSATION: "\texttt{\{INPUT\_DIALOGUE\}}"\textbackslash{}n
    
    SCORE:\\

    \hline
    \multicolumn{1}{|c|}{\textbf{Wikitactics}} \\
    \hline
    When there is a content accuracy dispute or a violation of Wikipedia’s neutral point of view policy, an editor can create a ‘dispute’ for a potentially problematic article, in which they provide their rationale, vote and discuss them with others. If the editors cannot reach an agreement, they can request mediation from a community volunteer, which is considered an escalation.\textbackslash{}n 
    
    Below, you have one dispute sourced from the Wikipedia Talk Page for which you have to predict escalation. We model this as a binary classification task by taking a textual dialogue as input to predict whether the dispute was eventually ‘escalated’ (y=1) or ‘non-escalated’ (y=0).\textbackslash{}n
                
    Important: Your prediction should consist of a single number between 0 and 1, with no additional text.\textbackslash{}n    
    
    DISPUTE: "\texttt{\{INPUT\_DIALOGUE\}}"\textbackslash{}n
    
    ESCALATION:\\

    \hline
    \multicolumn{1}{|c|}{\textbf{AFD}} \\
    \hline
    Below is a discussion between different individuals who discuss whether to delete a given Wikipedia article.\textbackslash{}n  
                
    The debate is initiated by an individual, who provides reasoning for nominating an article for deletion and then discusses it with other individuals. Administrators will in the end aggregate the discussion to decide whether to keep or delete the article.\textbackslash{}n  
    
    This is modelled as a binary classification task, taking an input discussion to predict if the outcome will be ‘Keep’ (y=1) or ‘Delete’ (y=0).\textbackslash{}n 
                
    Important: Your prediction should consist of a single number, with no additional text.\textbackslash{}n 
    
    DEBATE: "\texttt{\{INPUT\_DIALOGUE\}}"\textbackslash{}n
    
    OUTCOME:\\
    \hline
    \end{tabular}
    \caption{Prompts used to query 0-shot GPT-4o baselines across datasets. The prompts used for few-shot variants are identical but appending the few shot examples at the end of the prompt.}
    \label{tab:gpt4o_baseline_prompt}
\end{table*}

\subsection{Model, Budget and Hyperparameters}
\label{app:model_hyperparam}
The LLM feature-based models are based on the ridge regression and logistic regression algorithms. We tune the alpha parameter with values \([0.1, 1, 10, 100]\) and use the `auto' solver %(which can choose the solver automatically based on the type of data) 
for Ridge regression. We tune the regularisation parameter \(C\) with values \([0.1, 1, 10, 100]\) and solvers [`lbfgs', `liblinear', `sag', `saga'] for logistic regression. These models are fairly small, with the number of variables ranging from a few to 141 features (depending on which combinations of linguistic feature sets are included), except for the case when the features are disaggregated at the participant-level on the OUM dataset, which has 282 features. 

Longformer models are fine-tuned for 5 epochs with a batch size of 32 using the Adam optimiser, optimising: learning rates of 2e-5 and 1e-4, warmup steps corresponding to 1 and 3 epochs, and input sequence lengths of 2048 and 4096. The loss functions were Mean Absolute Error (MAE) and cross-entropy for regression and classification, respectively. These two variants of Longformers are fairly large, with around 435M parameters.

Table \ref{tab:best-found_hyperparameters} describes the best-found hyperparameter values. All other hyperparameters follow the default configurations. All the feature-based models are trained on an i5-14600K CPU, taking roughly 30 minutes in total across the three datasets. The Longformer models are trained on both NVIDIA RTX 3090 and Google Collab's A100 GPUs, depending on the GPU memory demand of the model, taking approximately 10 days for both hyperparameter-tuning and training.

\begin{table*}[h]
    \normalsize
    \centering
    \resizebox{\textwidth}{!}{
    \begin{tabular}{|l|p{4.6cm}|p{4.6cm}|p{4.6cm}|}
    \hline
        \textbf{Model\textbackslash Dataset} & \textbf{OUM} & \textbf{Wikitactics} & \textbf{AFD}\\
        \hline
        \multicolumn{4}{|c|}{\textbf{Neural Models}}\\
        \hline
        Longformer (full) & \#EoWU = 1, LR = 2e-5, MSL = 2048 & \#EoWU = 3, LR = 1e-4, MSL = 2048 & \#EoWU = 3, LR = 2e-5, MSL = 2048 \\
        Longformer (last layer) & \#EoWU = 1, LR = 1e-4, MSL = 4096 & \#EoWU = 1, LR = 1e-4, MSL = 4096 & \#EoWU = 3, LR = 1e-4, MSL = 4096 \\
        \hline
        \multicolumn{4}{|c|}{\textbf{(Standard) Feature-based Models}}\\
        \hline
        Politeness Markers & alpha=1 & C=1, solver=liblinear & C=10, solver=liblinear \\
        Collaboration Markers & alpha=0.1 & C=100, solver=saga & C=0.1, solver=lbfgs \\
        All Discrete Features & alpha=1 & C=10, solver=saga & C=10, solver=liblinear \\
        \hline
        \multicolumn{4}{|c|}{\textbf{LLM Feature-based Models}}\\
        \hline
        Dispute Tactics & alpha=0.1 & C=10, solver=liblinear & C=10, solver=liblinear \\
        QoA & alpha=10 & C=1, solver=lbfgs & C=1, solver=sag \\
        Information Content & alpha=0.1 & C=100, solver=saga & C=0.1, solver=lbfgs \\
        Style and Tone & alpha=1 & C=100, solver=sag & C=100, solver=lbfgs \\
        All LLM-generated Features & alpha=0.1 & C=1, solver=liblinear & C=1, solver=saga \\
        All Discrete \& LLM-generated Features & alpha=1 & C=0.1, solver=saga & C=100, solver=sag \\
        \hspace{0.2cm} \textit{$+$ participant-disaggregation} & alpha=10 & - & - \\
        \hline
    \end{tabular}
    }
    \caption{Best-found hyperparameters used for each model whose hyperparamters were tuned, across the three datasets. Acronyms used: \#EoWU=\#epochs of warm up; LR=learning rate; MSL=max sequence length.}
    \label{tab:best-found_hyperparameters}
\end{table*}

\subsection{XAI results}
\label{app:xai_analysis}
Table \ref{tab:XAI_coeffs} shows the model coefficients of the top-10 features with the highest FI to help interpret the model's decision-making and acquire insights into the relationship between distinct linguistic features and dialogue constructiveness.

Here we discuss the XAI analysis for the Wikitactics and AFD datasets, completing the subsection \ref{sec:XAI}. For Wikitactics, in predicting whether an edit would be `escalated' ($y=1$) or not ($y=0$), only three features have FIs $\geq$ 5\%. The first is the use of second-person pronouns (CM - you pronoun, $\bar{x}$; FI = 29.0\%, coeff = 0.475)%, a strong predictor of dispute escalation
. This corroborates the findings of \citep{zhang2018conversations, de2021beg}, reporting that the use of second-person pronouns is associated with unconstructive disagreements, as messages with `you' might be perceived as blameful \citep{gottman1989marital}.

Second, higher QoA (FI = 18.63\%, coeff = -0.194) corresponds to less escalation. By comparing many disputes with low and high QoA scores, we find that the explanations may be two manifolds: (i) disputes with high QoA usually involve individuals who behave professionally instead of throwing negative expressions, with the goal of reaching an agreement for the sake of improving the article, and (ii) well-supported arguments %from one side 
are more likely to garner support from other editors% on Wikipedia Talk Page
; this can be supported by the correlations between QoA with respect to `DT - Name calling/hostility, $\bar{x}$' ($r_s$ = -0.458), `DT - Ad hominem/ad argument, $\bar{x}$' ($r_s$ = -0.526), `DT - Providing clarification, $\bar{x}$' ($r_s$ = 0.363), and `ST - High politeness ($r_s$ = 0.431). %Thus, it can be concluded that these two phenomena promote resolution without needing moderator intervention. 

Third, hedging terms (CM - Hedging terms, $\bar{x}$; FI = 6.34\%, coeff = -0.172) have a negative association, suggesting expressions of uncertainty may reduce escalation. Upon inspecting dialogues with zero use versus frequent use of hedging terms, %the negative coefficient seems to be explainable for two reasons
it appears that (i) when individuals express uncertainty, they tend to be more willing to open the door for constructive discussion and negotiation, making it easier to find common ground, and (ii) if individuals do not express any uncertainty throughout the conversation, it is indicative that they are unlikely to compromise or reconsider their stance.

In the AFD dataset, when predicting whether an article would be `kept' ($y=1$) or `deleted' ($y=0$), the feature with the highest FI is `ST - Negative sentiment, $\bar{x}$' (FI = 35.89\%, coeff = -4.753). Likewise, `ST - Low politeness, $\bar{x}$' (FI = 7.85\%, coeff = -1.825) is associated with a higher likelihood of article deletion. These align with the intuition that extremely negative or impolite discussions are unproductive and may also indicate that a poorly written article %(that receives the request for deletion)
is more likely to attract negative sentiment. In contrast, positive sentiment words (PM - Positive sentiment words, $\bar{x}$; FI = 6.54\%, coeff = 2.231) and the adoption of a neutral sentiment (ST - Neutral sentiment, $\bar{x}$; FI = 14.83\%, coeff = 2.492) reduce the deletion likelihood.

\begin{table*}[h]
    \small
    \centering
    \begin{tabular}{|p{10.5cm}|p{2cm}|p{2cm}|}
    \hline
    \multicolumn{3}{|c|}{\textbf{OUM}} \\
    \hline
    \textbf{Feature} & \textbf{Coeff.} & \textbf{FI (\%)} \\
    \hline
    QoA & -0.203 & 13.79 \\
    PM - Hedge words, $\bar{x}$ & -1.731 & 10.76 \\
    CM - \# Headging terms, $\bar{x}$ & 0.735 & 10.37 \\
    DT - Contextualisation, $\bar{x}$ & -2.193 & 10.32 \\
    CM - \# 3rd person pronouns, $\bar{x}$ & 0.612 & 6.83 \\
    CM - \# me pronoun, $\bar{x}$ & 0.673 & 6.68 \\
    ST - Negative sentiment, $\bar{x}$ & -1.156 & 6.56 \\
    PM - Negative sentiment words, $\bar{x}$ & -1.383 & 6.18 \\
    CM - \# Geography terms, $\bar{x}$ & 0.649 & 5.58 \\
    PM - Positive sentiment words, $\bar{x}$ & 1.055 & 5.39 \\
    \hline
    \multicolumn{3}{|c|}{\textbf{Wikitactics}} \\
    \hline
    CM - \# you pronoun, $\bar{x}$ & 0.475 & 29.00 \\
    QoA & -0.194 & 18.63 \\
    CM - \# Headging terms, $\bar{x}$ & -0.172 & 6.34 \\
    ST - No uncertainty, $\bar{x}$ & 0.191 & 3.36 \\
    CM - \# me pronoun, $\bar{x}$ & 0.158 & 2.72 \\
    CM - \# Geography terms, $\bar{x}$ & -0.139 & 1.91 \\
    PM - 2nd person pronouns, $\bar{x}$ & 0.162 & 1.76 \\
    DT - Stating your stance/repeated argument, $\bar{x}$ & 0.057 & 1.75 \\
    DT - Providing clarification, $\bar{x}$ & -0.059 & 1.75 \\
    DT - Contextualisation, $\bar{x}$ & -0.103 & 1.61 \\
    \hline
    \multicolumn{3}{|c|}{\textbf{AFD}} \\
    \hline
    ST - Negative sentiment, $\bar{x}$ & -4.753 & 35.89 \\
    ST - Neutral sentiment, $\bar{x}$ & 2.492 & 14.83 \\
    ST - Low politeness, $\bar{x}$ & -1.825 & 7.85 \\
    PM - Positive sentiment words, $\bar{x}$ & 2.231 & 6.54 \\
    ST - No uncertainty, $\bar{x}$ & 1.485 & 3.83 \\
    DT - Providing clarification, $\bar{x}$ & 1.741 & 3.41 \\
    CM - \# words, $\nabla$ & 0.036 & 3.12 \\
    ST - Epistemic uncertainty, $\bar{x}$ & -1.335 & 2.82 \\
    ST - Negative sentiment, $\nabla$ & -1.401 & 2.43 \\
    DT - Stating your stance/repeated argument, $\bar{x}$ & -0.997 & 2.14 \\
    \hline
    \end{tabular}
    \caption{Model coefficients of the top-10 features with the highest FI. The following abbreviations are used to indicate the belongingness of feature sets: PM (Politeness Markers), CM (Collaboration Markers), DT (Dispute Tactics), and ST (Style and Tone).}
    \label{tab:XAI_coeffs}
\end{table*}

\subsection{License For Artifacts}

All datasets we included for our analysis were granted with permission from the original authors or are under a GNU general public license that allows free use for research purposes, i.e., consistent with their intended use (provided that it was specified).  The codes for using the proposed framework and reproducing the models we have trained, as well as all other assets produced in this work, are under the CC BY 4.0 license.

\end{document}